\documentclass[10pt,twocolumn,letterpaper]{article}

\usepackage[pagenumbers]{cvpr} 

\definecolor{cvprblue}{rgb}{0.21,0.49,0.74}
\usepackage[pagebackref,breaklinks,colorlinks,allcolors=cvprblue]{hyperref}

\def\confName{CVPR}
\def\confYear{2027}

\usepackage{hyperref}
\usepackage{url}

\usepackage{hyperref}
\usepackage{url}

\usepackage{amsmath}
\usepackage{booktabs}
\usepackage{enumitem}
\usepackage{hyperref}
\usepackage[capitalize, noabbrev]{cleveref}
\usepackage{subcaption}

\usepackage{algorithm}
\usepackage{algpseudocode}
\usepackage{pifont}        
\newcommand{\cmark}{\ding{51}}
\newcommand{\xmark}{\ding{55}}

\usepackage{graphicx}
\usepackage{amsmath}
\usepackage{cleveref}
\usepackage{multirow}
\usepackage{linguex}
\usepackage{sidecap}
\usepackage[utf8]{inputenc}
\usepackage[T1]{fontenc}

\usepackage{amssymb}
\usepackage{amsthm}

\usepackage{amsthm, amsmath, amssymb}
\newtheorem{proposition}{Proposition}

\usepackage[utf8]{inputenc} 
\usepackage[T1]{fontenc}    
\usepackage{hyperref}       
\usepackage{url}            
\usepackage{booktabs}       
\usepackage{amsfonts}       
\usepackage{nicefrac}       
\usepackage{microtype}      
\usepackage{xcolor}         

\usepackage[table,dvipsnames]{xcolor}

\usepackage{wrapfig}
\usepackage[most]{tcolorbox}
\tcbuselibrary{breakable}
\definecolor{promptbg}{RGB}{245, 248, 250}
\definecolor{promptborder}{RGB}{76, 175, 147}
\definecolor{prompttitle}{RGB}{76, 175, 147}

\usepackage{booktabs}
\usepackage[table]{xcolor}

\definecolor{bestgreen}{RGB}{232,245,235}

\title{ReJEPA: Bounded Predictive Regulation for Latent World Models}

\author{
\textbf{Fangzhou Lin}$^{1,2,3,5}$\quad
\textbf{Yuping Wang}$^{4}$\quad
\textbf{Shuo Xing}$^{1}$\quad
\textbf{Yang Chen}$^{5}$\quad
\textbf{Yuwen Heng}$^{5}$\quad\\[2pt]
\textbf{Xin Li}$^{1}$\footnotemark[2]\quad
\textbf{Wenbin He}$^{5}$\quad
\textbf{Yuhan Yao}$^{5}$\quad
\textbf{Zhengzhong Tu}$^{1}$\footnotemark[2]\quad
\textbf{Liu Ren}$^{5}$\footnotemark[1]\\[4pt]
\footnotesize
$^{1}$ Texas A\&M University\;
$^{2}$ Worcester Polytechnic Institute\;
$^{3}$Tohoku University \\
\footnotesize
$^{4}$University of Michigan  \;
\footnotesize
$^{5}$Bosch Research North America \& Bosch Center for AI \;\\
\footnotesize
}

\begin{document}
\def\customfootnotetext#1#2{{%
  \let\thefootnote\relax
  \footnotetext[#1]{#2}}}

\maketitle

\customfootnotetext{2}
{\textsuperscript{$\dagger$}Advising: Xin Li and Zhengzhong Tu}
\customfootnotetext{1}{\textsuperscript{*}Corresponding Author: Liu Ren}

\begin{abstract}
Latent world models learn representations by predicting future states in an abstract embedding space rather than reconstructing raw pixels. When training across multi-frame sequences, however, not all frames and spatial regions are equally informative or predictable. We argue that forcing a world model to minimize every prediction error may be unnecessary and harmful. Under standard $L_2$ loss, unbounded gradients force the encoder and predictor to overfit noisy or transient frames at the expense of core physical structure, while amplifying gradient conflict with representation regularizers. Conversely, $L_1$ loss bounds large gradients, but its constant gradient magnitude near zero disrupts the local smoothness required for planning. We resolve this tension with \textbf{ReJEPA}, a bounded predictive regulation principle that combines bounded gradients for large errors with smooth quadratic behavior near zero, instantiated simply via the element-wise $\log\cosh$ loss.
 As a drop-in training replacement across Temporal Straightening, LeWorldModel, and V-JEPA~2.1, $\log\cosh$ consistently outperforms $L_2$, a curvature-matched quadratic control, $L_1$, and Smooth-$L_1$ baselines without altering architectures or inference procedures. Notably, the gains grow with environmental complexity, scaling from simple navigation to contact-rich manipulation and real-world autonomous driving on NAVSIM. Probing and geometric analyses confirm that relaxing the penalty on extreme errors yields representations that better capture physical states and align with the true planning cost landscape.
\end{abstract}    
\section{Introduction}
\label{sec:introduction}

\begin{figure}[t]
    \vspace{-0.1in}
    \centering
    \includegraphics[width=\linewidth]{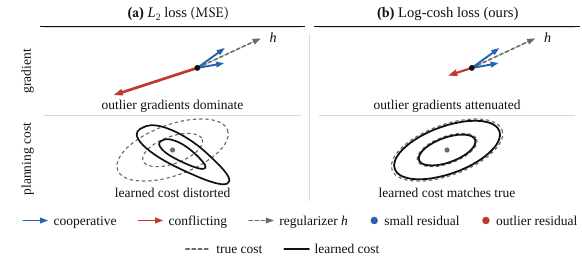}
    \vspace{-0.26in}
\caption{\textbf{Effect of bounded predictive regulation on training gradients and planning geometry.} \textbf{(a)} Under $L_2$ loss, large outlier residuals (red) dominate parameter updates against the regularizer gradient $h$ (dashed gray) and distort the learned planning cost contours (solid black) away from the true cost (dashed gray). \textbf{(b)} Our $\log\cosh$ penalty attenuates outlier gradients while preserving small cooperative residuals (blue), aligning the learned planning cost landscape with the true system.}

    \label{fig:predictive-influence}
    \vspace{-0.5cm}
\end{figure}

\noindent\textbf{Latent World Models.}
A central premise of latent world models is that effective decision-making does not require reconstructing every pixel of the environment~\citep{lecun2022path,kappiyath2026self,zhou2026hierarchical,balestriero2026proving,yalovetzky2026lehami,toso2026preserving,zhang2026h,zhu2026ad,goswami2026world}. Instead, these models encode high-dimensional observations into compact latent states and learn predictive dynamics directly in representation space~\citep{assran2023self,garrido2024learning,bardes2024revisiting,assran2025v,mur2026v}. Their downstream utility depends on two coupled properties: the physical state information retained by the encoder, and how the latent geometry guides action search. In goal-conditioned planning, a predictor rolls out future latent states under candidate actions, and a planner selects actions whose predicted outcomes lie close to the goal~\citep{pertsch2020long,wang2023describe,zhang2026world}. Existing methods learn these representations by adapting pretrained features or training encoders end-to-end~\citep{zhou2024dino,destrade2025value,balestriero2025lejepa}, often adding explicit structural regularizers to straighten trajectories or prevent collapse~\citep{wang2026ts,maes2026leworldmodel,kuang2026lpwm}. Despite architectural differences, all of these approaches rely on minimizing prediction errors across multi-frame sequences to shape the latent space.

\noindent\textbf{Why Minimizing Every Prediction Error Might Hurt.}
Predictive models learn by penalizing discrepancies between predicted and target latent states~\citep{doerr2018probabilistic,henaff2019model,lecun2022path}. Across multi-frame sequences, however, not all frames and spatial regions are equally clean or task-relevant. Complex control trajectories and real-world videos naturally contain abrupt contact transitions, uninformative intermediate frames, occlusions, and background clutter. Unlike supervised regression toward fixed labels, the prediction targets in a latent world model are generated by an evolving encoder~\citep{gidaris2017detect,muller2019does}. Forcing every prediction error to zero is therefore unnecessary and often harmful. Under standard $L_2$ loss, the derivative grows linearly with the residual~\citep{efron1991regression,chen2018robust,hu2025beyond}, allowing large errors from noisy or transient frames to dominate parameter updates. To accommodate these outliers, the encoder can distort its latent geometry or discard subtle physical states critical for control. Conversely, while $L_1$ loss bounds large gradients, its derivative maintains a constant unit magnitude even for arbitrarily small errors. This lack of local smoothness near zero induces oscillation around well-predicted states and degrades the local cost curvature required for planning.

\noindent\textbf{Bounded Predictive Regulation and Representation Shaping.}
These observations point to a simple design principle: a predictive objective should bound gradient magnitudes for large errors so the model does not overfit noisy transitions, while remaining smooth and quadratic near zero to support fine-grained planning. Bounding large residuals also benefits models trained with explicit representation regularizers, such as trajectory straightening in Temporal Straightening (TS)~\citep{wang2026ts} or distributional anti-collapse regularization in LeWorldModel (LeWM)~\citep{maes2026leworldmodel}. Because predictive and structural objectives update shared parameters, large prediction errors under $L_2$ loss can severely amplify gradients that conflict with the regularizer. Through a residual-space decomposition, we show that attenuating large residuals improves local gradient compatibility whenever the suppressed outliers carry more conflict than cooperation.

\noindent\textbf{Our Approach and Empirical Findings.}
We term this framework \textbf{ReJEPA} (bounded predictive regulation for JEPA-style latent world models), and instantiate it simply with the hyperparameter-free, element-wise $\log\cosh$ loss. Without introducing threshold hyperparameters, $\log\cosh$ behaves quadratically near zero and smoothly saturates its derivative to $(-1, 1)$ for large errors. We apply $\log\cosh$ as a drop-in training replacement across TS~\citep{wang2026ts}, LeWM~\citep{maes2026leworldmodel}, and V-JEPA~2.1~\citep{mur2026v} while keeping all network architectures, regularizer weights, and inference-time planning procedures unchanged. To verify that the gains stem from nonlinear large-error attenuation rather than global loss rescaling, we compare $\log\cosh$ against a locally curvature-matched $L_2$ control alongside standard $L_2$, $L_1$, and Smooth-$L_1$ baselines.

Across benchmarks, $\log\cosh$ consistently improves over all baselines, with gains that scale noticeably with environmental complexity and observation noise. On simple navigation tasks with clean dynamics (Wall and TwoRoom), $\log\cosh$ yields steady improvements over quadratic baselines. On harder simulated tasks involving contact dynamics (PushT, Cube), multi-corridor layouts (PointMaze), and longer planning horizons, the gains expand up to $12$ percentage points in open-loop planning success. When scaled to real-world multi-frame driving in NAVSIM~\citep{dauner2024navsim}, where background clutter and dynamic traffic introduce severe prediction uncertainty, $\log\cosh$ lowers perception and trajectory errors by $9.4\%$--$17.6\%$ and improves end-to-end closed-loop driving performance. Furthermore, frozen-encoder probing on PushT (reducing state MSE by over $20\%$) and action-space visualizations confirm that relaxing the penalty on extreme errors enables the latent space to better capture physical states and align with the true planning cost landscape.

We summarize our main contributions as follows:
\begin{itemize}[nosep, leftmargin=*]
    \item We identify the pitfall of unconstrained error minimization across multi-frame sequences in latent world models, showing that bounding large predictive gradients while preserving smooth origin curvature prevents noisy frames from distorting the representation.
    \item We instantiate this principle with the hyperparameter-free $\log\cosh$ loss, compare its weighting properties against $L_2$, curvature-matched $L_2$, $L_1$, and Smooth-$L_1$ losses, and characterize when residual attenuation reduces gradient conflict with structural regularizers.
    \item We validate our approach across TS, LeWM, and V-JEPA~2.1 from simulated control to real-world autonomous driving, demonstrating gains that scale with data complexity alongside improved physical state accessibility and planning geometry.
\end{itemize}
\section{Related Work}
\label{sec:related_work}

\noindent\textbf{Latent World Models and Predictive Representations.}
Learning predictive dynamics in compact latent spaces enables efficient planning without pixel-level reconstruction~\citep{ha2018world,hafner2019learning,lecun2022path}. Model-based reinforcement learning methods such as Dreamer and TD-MPC learn latent transition models alongside reward or value objectives for trajectory optimization~\citep{hafner2020dream,hansen2022temporal}. In self-supervised learning, Joint-Embedding Predictive Architectures (JEPA) and non-contrastive frameworks predict target representations directly from context views~\citep{grill2020bootstrap,chen2021exploring,assran2023self,bardes2024revisiting}. Extending this paradigm to action-conditioned control, recent world models either train dynamics predictors atop frozen visual backbones~\citep{zhou2025dinowm,sobal2025planning} or optimize encoders and predictors jointly across multi-frame sequences~\citep{assran2025v,maes2026leworldmodel}.

\noindent\textbf{Structural Regularization of Latent Spaces.}
Optimizing representation prediction alone admits trivial collapsed solutions unless structural constraints are imposed. Contrastive methods prevent collapse by balancing feature alignment and uniform distribution on the hypersphere~\citep{wang2020understanding}. Non-contrastive approaches enforce explicit statistical or geometric priors, including redundancy reduction in Barlow Twins~\citep{zbontar2021barlow}, variance-covariance regularization in VICReg~\citep{bardes2022vicreg}, and isotropic Gaussian regularization in LeWM~\citep{maes2026leworldmodel}. Beyond distributional spread, recent world models constrain latent dynamics via temporal trajectory straightening~\citep{wang2026ts} or sparse coding~\citep{kuang2026lpwm}.

\noindent\textbf{Robust Penalties and Multi-Objective Gradients.}
Bounded-derivative penalties are foundational in robust statistics and M-estimation to limit the influence of outlier residuals~\citep{hampel1986robust,huber1992robust,charbonnier1997deterministic,barron2019general}. In deep learning, piecewise Smooth-$L_1$ losses are widely used for regression stability~\citep{girshick2015fast}, and $L_1$ penalties have been adopted in video predictive architectures to stabilize training~\citep{bardes2024revisiting}. When multiple objectives update shared parameters, multi-task optimization methods such as MGDA~\citep{sener2018multi}, GradNorm~\citep{chen2018gradnorm}, PCGrad~\citep{yu2020gradient}, and CAGrad~\citep{liu2021conflict} explicitly balance gradient norms or project away conflicting gradient directions.

\section{Bounded Predictive Regulation in Latent World Models}
\label{sec:method}

We first formalize the optimization structure of latent world models across multi-frame sequences. We then analyze predictive penalties through their coordinate-wise gradient weights, compare unbounded, non-smooth, piecewise, and smoothly bounded formulations, and characterize how bounding large residuals interacts with explicit representation regularizers.

\subsection{Multi-Frame Prediction and Residual Weighting}
\label{sec:method:setup}

Consider a latent world model operating on multi-frame observation sequences $\{o_t\}_{t=1}^T$, optionally conditioned on actions or spatiotemporal masks. A context encoder $f_\theta$ maps visible observations to latent states $Z = f_\theta(o)$. A predictor $g_\phi$ maps $Z$ (and any conditioning variables) to predicted latent states $\hat Z = g_\phi(Z)$ for future frames or masked regions. Target latent states $Z^+ = f_\xi(o^+)$ are produced by a target encoder $f_\xi$, where both $\hat Z$ and $Z^+$ include each model's native pre-loss normalization (such as LayerNorm). Across model families, $f_\xi$ may share weights with $f_\theta$ end-to-end ($\xi = \theta$, as in LeWM~\citep{maes2026leworldmodel}), apply a stop-gradient operator ($\xi = \operatorname{sg}(\theta)$, as in TS~\citep{wang2026ts}), or update via an exponential moving average of $\theta$ (as in V-JEPA~2.1~\citep{mur2026v}). Let $\vartheta = (\theta, \phi)$ denote all trainable parameters updated by the training objective.

Let $r(\vartheta) = \operatorname{vec}(\hat Z - Z^+) \in \mathbb R^m$ be the flattened vector of prediction residuals across all predicted frames, spatial tokens, and feature channels. Each coordinate $r_i$ ($i \in \{1, \dots, m\}$) measures the prediction error at a specific frame, spatial location, and channel. We consider training objectives of the form
\begin{equation}
\mathcal L_\rho(\vartheta)
=
\underbrace{c\sum_{i=1}^{m}\rho(r_i(\vartheta))}_{\mathcal P_\rho(\vartheta)}
+
\underbrace{\lambda\mathcal R(\vartheta)}_{\mathcal G(\vartheta)},
\label{eq:world-model-objective}
\end{equation}
where $\rho: \mathbb R \to \mathbb R_{\ge 0}$ is a symmetric scalar penalty applied element-wise, and $c > 0$ is a parameter-independent reduction factor (or coordinate-specific positive weight in multi-branch masked models). The term $\mathcal G(\vartheta) = \lambda\mathcal R(\vartheta)$ is an explicit structural regularizer weighted by $\lambda \ge 0$, such as trajectory straightening in TS~\citep{wang2026ts} or anti-collapse regularization in LeWM~\citep{maes2026leworldmodel} (with $\mathcal G(\vartheta) = 0$ in models that rely solely on momentum targets and predictive branches). Within each model family, we vary only $\rho$ while holding the architecture, regularizer $\mathcal R$, reduction factors, weight $\lambda$, and schedules fixed.

Because both the predictor and context encoder depend on $\vartheta$, minimizing $\mathcal P_\rho(\vartheta)$ shapes the representation alongside the dynamics model. At a fixed parameter state $\vartheta$ and minibatch realization (holding stopped target paths constant while differentiating through active encoder, predictor, and normalization paths), the predictive gradient on $\vartheta$ is
\begin{equation}
p_\rho
\equiv\nabla_\vartheta\mathcal P_\rho(\vartheta)
=
c\sum_{i=1}^{m}\psi_\rho(r_i)\nabla_\vartheta r_i,
\label{eq:predictive-gradient}
\end{equation}
where $\psi_\rho(u) = \rho'(u)$ is the derivative (residual score) at scalar input $u$. Each residual $r_i$ contributes $c\psi_\rho(r_i)\nabla_\vartheta r_i$ to the update. Because gradient directions $\nabla_\vartheta r_i$ vary across coordinates, their relative scores $\psi_\rho(r_i)$ dictate both the magnitude and net direction of $p_\rho$.

Across multi-frame sequences, individual frames and spatial regions differ widely in predictability and task relevance. Abrupt contacts, occlusions, and background clutter naturally yield large residuals $|r_i|$. Under standard squared error $\rho_{\mathrm{mse}}(u) = u^2$, the score $\psi_{\mathrm{mse}}(u) = 2u$ grows linearly without bound, allowing outlier frames to dominate parameter updates. Because target states $Z^+$ depend on the evolving encoder, forcing the model to aggressively minimize large errors on noisy frames can distort the latent geometry or discard subtle physical states critical for control.

\subsection{Dual-Regime Weighting via Log-Cosh}
\label{sec:method:influence}
\label{sec:method:instantiation}

Preventing outlier transitions from dominating updates requires bounding $\psi_\rho(u)$ for large residuals. However, bounding large gradients alone is insufficient. Under absolute error ($L_1$, $\rho_{\ell_1}(u) = |u|$), the score $\psi_{\ell_1}(u) = \operatorname{sign}(u)$ is bounded to $\{\pm 1\}$, but its magnitude remains constant down to zero. When a residual $r_i$ is already small (on well-predicted states), $L_1$ still applies unit-magnitude gradient steps, inducing oscillation around zero and degrading the smooth local geometry needed by planners. Piecewise penalties such as Smooth-$L_1$ (Huber loss with threshold $\beta > 0$) combine quadratic and linear regimes, but require a scale-sensitive threshold $\beta$ and have a discontinuous second derivative at $|u| = \beta$.

To achieve both bounded sensitivity to large errors and smooth local curvature without threshold hyperparameters, we use the element-wise $\log\cosh$ penalty:
\begin{equation}
\mathcal L_{\mathrm{lc}}(\vartheta)
=
c\sum_{i=1}^{m}\log\cosh(r_i(\vartheta))+\lambda\mathcal R(\vartheta).
\label{eq:final-objective}
\end{equation}
For $u \in \mathbb R$, $\log\cosh(u)$ is infinitely differentiable ($C^\infty$). Near zero, $\log\cosh(u) = \tfrac12 u^2 - \tfrac1{12}u^4 + O(u^6)$, matching quadratic loss. For large $|u|$, it approaches $|u| - \log 2$. Its first derivative $\psi_{\mathrm{lc}}(u) = \tanh(u)$ is linear with unit slope near zero and saturates smoothly to $(-1, 1)$ as $|u| \to \infty$. Its second derivative $\rho_{\mathrm{lc}}''(u) = \operatorname{sech}^2(u)$ decays smoothly from $1$ at $u=0$ toward $0$ for large errors.

\begin{table}[h]
\centering
\caption{
Comparison of scalar predictive penalties $\rho(u)$, derivatives $\psi_\rho(u) = \rho'(u)$, residual weights $w_\rho(u) = \psi_\rho(u)/u$ (with $w_\rho(0) = \rho''(0)$), origin curvature $\rho''(0)$, gradient bound $\sup_u |\psi_\rho(u)|$, and smoothness class. Smooth-$L_1$ denotes the unit-threshold Huber loss ($\beta=1$).
}
\vspace{-2mm}
\label{tab:predictive-penalties}
\setlength{\tabcolsep}{3.5pt}
\renewcommand{\arraystretch}{0.9}
\resizebox{\columnwidth}{!}{%
\begin{tabular}{lcccccc}
\toprule
Penalty & $\rho(u)$ & $\psi_\rho(u)$ & $w_\rho(u)$ & $\rho''(0)$ & $\sup|\psi_\rho|$ & Class \\
\midrule
Quadratic ($L_2$) & $u^2$ & $2u$ & $2$ & $2$ & $\infty$ & $C^\infty$ \\
Matched $L_2$ & $\tfrac12u^2$ & $u$ & $1$ & $1$ & $\infty$ & $C^\infty$ \\
Absolute ($L_1$) & $|u|$ & $\operatorname{sign}(u)$ & $1/|u|$ & undef. & $1$ & $C^0$ \\
Smooth-$L_1$ & $\begin{cases}\tfrac12 u^2, |u|\le 1\\ |u|-\tfrac12, |u|>1\end{cases}$ & $\operatorname{clip}(u,-1,1)$ & $\min(1,\tfrac{1}{|u|})$ & $1$ & $1$ & $C^1$ \\
Log-cosh & $\log\cosh(u)$ & $\tanh(u)$ & $\tanh(u)/u$ & $1$ & $1$ & $C^\infty$ \\
\bottomrule
\end{tabular}%
}
\vspace{-3mm}
\end{table}

\noindent\textbf{Predictive Influence Operator.}
Let $J = D_\vartheta r \in \mathbb R^{m \times d}$ be the Jacobian of $r$ with respect to $\vartheta \in \mathbb R^d$. Defining the weight function $w_\rho(u) = \psi_\rho(u)/u$ for $u \neq 0$ (with $w_\rho(0) = \rho''(0)$ when twice differentiable at zero) yields a diagonal \emph{Predictive Influence Operator}:
\begin{equation}
\mathsf W_\rho(r)=\operatorname{diag}\!\bigl(w_\rho(r_1),\ldots,w_\rho(r_m)\bigr), \quad
p_\rho=cJ^\top\mathsf W_\rho(r)r.
\label{eq:influence-pullback}
\end{equation}
Table~\ref{tab:predictive-penalties} compares $\mathsf W_\rho(r)$ across penalties. Under $u^2$ and $\tfrac12 u^2$, $\mathsf W_\rho(r)$ is a constant multiple of the identity ($2I$ and $I$), weighting clean and noisy frames equally. Under $L_1$, $w_{\ell_1}(u) = 1/|u|$ diverges as $|u| \to 0$, over-weighting tiny residuals. Under Smooth-$L_1$, $w(u)$ has a sharp kink at $|u|=1$. Under $\log\cosh$, $w_{\mathrm{lc}}(r_i) = \tanh(r_i)/r_i \in (0, 1]$ decreases smoothly with $|r_i|$. Because this attenuation varies across coordinates, $\mathsf W_{\mathrm{lc}}(r)$ changes both the norm and direction of $\mathsf W_{\mathrm{lc}}(r)r$ before $J^\top$ maps it to parameter space.

\noindent\textbf{Locally Curvature-Matched Quadratic Control.}
Replacing $u^2$ with $\log\cosh(u)$ changes both the origin curvature (from $2$ to $1$) and large-error saturation. To separate these effects, we test the matched quadratic control $\rho_{\mathrm{q}}(u) = \tfrac12 u^2$, which shares unit origin curvature ($\rho_{\mathrm{q}}''(0) = \rho_{\mathrm{lc}}''(0) = 1$) and satisfies $p_{\mathrm{q}} = \tfrac12 p_{\mathrm{mse}}$. When $\mathcal G = 0$, halving the loss scales raw gradients uniformly, which is largely canceled by second-moment normalization in AdamW (up to $\epsilon$, weight decay, and gradient clipping). When $\mathcal G \neq 0$, $\tfrac12 u^2$ uniformly halves the weight of prediction relative to $\mathcal G$. In contrast, $\log\cosh$ applies coordinate-wise weights $w_{\mathrm{lc}}(r_i) \in (0, 1]$, changing the relative direction of $p_\rho$ across frames and channels rather than applying a global scalar factor.

\subsection{Interaction with Representation Regularizers}
\label{sec:method:interaction}

In models with an explicit regularizer $\mathcal G(\vartheta) = \lambda\mathcal R(\vartheta)$ ($\lambda > 0$), such as TS~\citep{wang2026ts} or LeWM~\citep{maes2026leworldmodel}, large prediction residuals from noisy frames can also conflict with $\mathcal G$. Let $h = \nabla_\vartheta\mathcal G(\vartheta)$ be the regularizer gradient on $\vartheta$ and define $A_\rho = \langle p_\rho, h \rangle$. Under a local raw-gradient step $\vartheta^+ = \vartheta - \eta(p_\rho + h)$ with $\eta > 0$, a first-order expansion of differentiable $\mathcal G$ as $\eta \to 0^+$ gives
\begin{equation}
\mathcal G(\vartheta^+)-\mathcal G(\vartheta)
=
-\eta\bigl(\|h\|^2+A_\rho\bigr)+o(\eta).
\label{eq:representation-progress}
\end{equation}
The predictive update reinforces $\mathcal G$ when $A_\rho > 0$ (cooperation) and opposes it when $A_\rho < 0$ (conflict). Writing $s = J h \in \mathbb R^m$, where $s_i = \langle \nabla_\vartheta r_i, h \rangle$, decomposes $A_\rho$ in residual space:
\begin{equation}
A_\rho
=
c\langle\psi_\rho(r),s\rangle
=
c\sum_{i=1}^{m}\psi_\rho(r_i)s_i.
\label{eq:interaction-pullback}
\end{equation}
For the matched quadratic $\tfrac12 u^2$, $\psi_{\mathrm{q}}(r_i) = r_i$, so $A_{\mathrm{q}} = c\sum_{i=1}^m \alpha_i$ with $\alpha_i = r_i s_i$. Coordinate $i$ cooperates with $\mathcal G$ when $\alpha_i > 0$ and conflicts when $\alpha_i < 0$.

\noindent\textbf{When Residual Attenuation Reduces Conflict.}
Relative to $\tfrac12 u^2$, $\log\cosh$ scales coordinate $i$ by $w_{\mathrm{lc}}(r_i) = 1 - \delta_i$, where $\delta_i = 1 - \tanh(r_i)/r_i \in [0, 1)$ is the fractional attenuation ($\delta_i = 0$ at $r_i = 0$).

\begin{proposition}[Residual-space interaction]
\label{prop:residual-interaction}
At the same state $\vartheta$ and minibatch realization, $\log\cosh$ and the matched quadratic $\tfrac12 u^2$ satisfy
\begin{equation}
\begin{aligned}
A_{\mathrm{lc}}-A_{\mathrm q}
=c\Bigg[
\underbrace{
\sum_{i:\,\alpha_i<0}\delta_i|\alpha_i|
}_{\text{removed conflict}}
-\underbrace{
\sum_{i:\,\alpha_i>0}\delta_i\alpha_i
}_{\text{lost cooperation}}
\Bigg].
\end{aligned}
\label{eq:residual-interaction}
\end{equation}
Thus, $A_{\mathrm{lc}} > A_{\mathrm{q}}$ if and only if the removed conflict exceeds the lost cooperation.
\end{proposition}

\begin{proof}
Since $\psi_{\mathrm{lc}}(r_i) = (1 - \delta_i)r_i$, Equation~\ref{eq:interaction-pullback} gives $A_{\mathrm{lc}} = c\sum_{i=1}^m (1 - \delta_i)\alpha_i$. Subtracting $A_{\mathrm{q}} = c\sum_{i=1}^m \alpha_i$ and separating indices by the sign of $\alpha_i$ yields Equation~\ref{eq:residual-interaction}.
\end{proof}

Because $\delta_i = \tfrac13 r_i^2 + O(r_i^4)$ near zero and $\delta_i \to 1$ as $|r_i| \to \infty$, Equation~\ref{eq:residual-interaction} is dominated by large residuals. When outlier errors from noisy transitions pull the encoder away from smooth or well-spread representations ($\alpha_i < 0$), attenuating them removes substantial conflict while sacrificing very little cooperation on well-predicted states (where $\delta_i \approx 0$).

\subsection{Applying the Replacement}
\label{sec:method:implementation}

We apply $\log\cosh$ as a unit-scale training replacement for $\rho$ (evaluated stably via Equation~\ref{eq:app-analytic-stable-logcosh} in Appendix~\ref{app:analytical-details}), preserving each model's native normalization, masking, reductions, and target rules. In TS and LeWM, architectures, regularizers, weights $\lambda$, and planners remain unchanged. In our NAVSIM driving models, $\rho$ replaces the feature regression penalty across both the two-branch convolutional JEPA and the 4-layer hierarchical V-JEPA~2.1 ViT-L architecture (Appendix~\ref{app:driving-details}). At inference time, all goal-conditioned planners retain the original squared Euclidean distance in latent space as the planning cost.

\providecommand{\expstat}[2]{#1\,{\scriptsize$\pm$\,#2}}
\providecommand{\tbd}[2]{#1\,{\scriptsize$\pm$\,#2}}

\section{Experiments}
\label{sec:experiments}

\begin{table*}[h]
\centering
\caption{Goal-reaching success rates (\%) for Temporal Straightening (TS) using the gradient descent (GD) planner. Values are mean $\pm$ standard deviation over three data-sampling seeds, with 50 test episodes per seed. Best means per group are in \textbf{bold}.}
\vspace{-2mm}
\label{tab:main_results}
\setlength{\tabcolsep}{5.2pt}
\renewcommand{\arraystretch}{0.9}
\resizebox{\textwidth}{!}{%
\begin{tabular}{llc|cccccccc}
\toprule
\multirow{2}{*}{Encoder} & \multirow{2}{*}{Shape}
& \multirow{2}{*}{Penalty}
& \multicolumn{2}{c}{Wall} & \multicolumn{2}{c}{PointMaze--UMaze}
& \multicolumn{2}{c}{PointMaze--Medium} & \multicolumn{2}{c}{PushT} \\
\cmidrule(lr){4-5}\cmidrule(lr){6-7}\cmidrule(lr){8-9}\cmidrule(lr){10-11}
& & & Open-loop & MPC & Open-loop & MPC & Open-loop & MPC & Open-loop & MPC \\
\midrule
\multirow{5}{*}{\shortstack[l]{DINOv2\\(patch) + proj}} & \multirow{5}{*}{$1\times384$}
& $u^2$ & \expstat{36.67}{3.40} & \expstat{51.33}{4.11} & \expstat{30.67}{3.40} & \expstat{90.00}{3.27} & \expstat{16.67}{4.11} & \expstat{69.33}{3.40} & \expstat{2.67}{2.49} & \expstat{10.00}{1.63} \\
& & $\frac12u^2$ & \expstat{38.67}{2.49} & \expstat{54.00}{3.27} & \expstat{34.00}{3.27} & \expstat{92.67}{1.89} & \expstat{19.33}{4.11} & \expstat{73.33}{2.49} & \expstat{4.00}{1.63} & \expstat{12.00}{1.63} \\
& & $L_1$ ($|u|$) & \expstat{40.67}{3.40} & \expstat{53.33}{4.99} & \expstat{36.00}{4.32} & \expstat{93.33}{2.49} & \expstat{18.67}{4.99} & \expstat{76.00}{3.27} & \expstat{4.67}{2.49} & \expstat{12.67}{2.49} \\
& & Smooth-$L_1$ & \expstat{39.33}{2.49} & \expstat{55.33}{3.40} & \expstat{35.33}{2.49} & \expstat{94.67}{1.89} & \expstat{21.33}{4.11} & \expstat{74.67}{2.49} & \expstat{4.00}{1.63} & \expstat{13.33}{1.89} \\
& & Log-cosh & \expstat{\textbf{42.00}}{3.27} & \expstat{\textbf{56.67}}{4.11} & \expstat{\textbf{38.67}}{3.40} & \expstat{\textbf{96.00}}{0.00} & \expstat{\textbf{22.67}}{5.73} & \expstat{\textbf{78.00}}{2.83} & \expstat{\textbf{5.33}}{3.40} & \expstat{\textbf{14.67}}{0.94} \\
\addlinespace[2pt]
\multirow{5}{*}{\shortstack[l]{ResNet\\(from scratch)}} & \multirow{5}{*}{$1\times384$}
& $u^2$ & \expstat{78.67}{7.36} & \expstat{98.00}{1.63} & \expstat{42.00}{5.89} & \expstat{80.00}{3.27} & \expstat{44.67}{6.18} & \expstat{94.00}{1.63} & \expstat{14.67}{4.11} & \expstat{42.00}{6.53} \\
& & $\frac12u^2$ & \expstat{81.33}{5.73} & \expstat{99.33}{0.94} & \expstat{45.33}{5.73} & \expstat{83.33}{0.94} & \expstat{48.00}{6.53} & \expstat{96.00}{1.63} & \expstat{16.67}{4.11} & \expstat{45.33}{5.73} \\
& & $L_1$ ($|u|$) & \expstat{82.67}{6.60} & \expstat{98.67}{1.89} & \expstat{46.67}{7.36} & \expstat{85.33}{2.49} & \expstat{51.33}{6.60} & \expstat{95.33}{2.49} & \expstat{16.00}{4.32} & \expstat{47.33}{5.73} \\
& & Smooth-$L_1$ & \expstat{80.67}{5.73} & \expstat{99.33}{0.94} & \expstat{49.33}{5.73} & \expstat{84.00}{1.63} & \expstat{50.00}{5.89} & \expstat{96.67}{0.94} & \expstat{18.00}{3.27} & \expstat{46.00}{4.90} \\
& & Log-cosh & \expstat{\textbf{84.00}}{7.12} & \expstat{\textbf{100.00}}{0.00} & \expstat{\textbf{52.00}}{6.53} & \expstat{\textbf{86.67}}{0.94} & \expstat{\textbf{54.00}}{7.12} & \expstat{\textbf{98.00}}{0.00} & \expstat{\textbf{19.33}}{3.40} & \expstat{\textbf{48.67}}{4.99} \\
\addlinespace[2pt]
\multirow{5}{*}{\shortstack[l]{DINOv2\\(patch) + proj}} & \multirow{5}{*}{$14^2\times8$}
& $u^2$ & \expstat{86.00}{3.27} & \expstat{98.00}{1.63} & \expstat{82.00}{4.32} & \expstat{95.33}{2.49} & \expstat{72.00}{4.32} & \expstat{94.67}{2.49} & \expstat{78.00}{4.32} & \expstat{80.67}{2.49} \\
& & $\frac12u^2$ & \expstat{87.33}{2.49} & \expstat{99.33}{0.94} & \expstat{84.00}{1.63} & \expstat{98.00}{1.63} & \expstat{74.00}{4.32} & \expstat{96.67}{1.89} & \expstat{79.33}{3.40} & \expstat{82.67}{3.77} \\
& & $L_1$ ($|u|$) & \expstat{86.67}{3.40} & \expstat{99.33}{0.94} & \expstat{89.33}{3.40} & \expstat{97.33}{2.49} & \expstat{76.67}{4.99} & \expstat{97.33}{1.89} & \expstat{80.67}{4.11} & \expstat{82.00}{4.32} \\
& & Smooth-$L_1$ & \expstat{88.67}{2.49} & \expstat{98.67}{1.89} & \expstat{88.00}{2.83} & \expstat{98.67}{0.94} & \expstat{80.00}{3.27} & \expstat{96.00}{1.63} & \expstat{78.67}{3.40} & \expstat{84.00}{3.27} \\
& & Log-cosh & \expstat{\textbf{90.00}}{1.63} & \expstat{\textbf{100.00}}{0.00} & \expstat{\textbf{94.00}}{1.63} & \expstat{\textbf{100.00}}{0.00} & \expstat{\textbf{84.00}}{3.27} & \expstat{\textbf{98.67}}{0.94} & \expstat{\textbf{82.00}}{4.32} & \expstat{\textbf{85.33}}{4.99} \\
\addlinespace[2pt]
\multirow{5}{*}{\shortstack[l]{ResNet\\(from scratch)}} & \multirow{5}{*}{$14^2\times8$}
& $u^2$ & \expstat{80.00}{3.27} & \expstat{98.67}{0.94} & \expstat{53.33}{6.60} & \expstat{94.67}{2.49} & \expstat{69.33}{4.11} & \expstat{96.00}{1.63} & \expstat{66.00}{3.27} & \expstat{84.00}{4.32} \\
& & $\frac12u^2$ & \expstat{82.00}{3.27} & \expstat{\textbf{100.00}}{0.00} & \expstat{57.33}{6.60} & \expstat{96.00}{1.63} & \expstat{73.33}{2.49} & \expstat{97.33}{0.94} & \expstat{68.00}{2.83} & \expstat{87.33}{3.77} \\
& & $L_1$ ($|u|$) & \expstat{83.33}{3.40} & \expstat{98.67}{1.89} & \expstat{58.67}{7.36} & \expstat{97.33}{1.89} & \expstat{77.33}{3.40} & \expstat{96.67}{1.89} & \expstat{67.33}{3.40} & \expstat{89.33}{4.11} \\
& & Smooth-$L_1$ & \expstat{81.33}{2.49} & \expstat{99.33}{0.94} & \expstat{61.33}{6.60} & \expstat{96.67}{2.49} & \expstat{75.33}{2.49} & \expstat{98.00}{1.63} & \expstat{69.33}{2.49} & \expstat{88.00}{3.27} \\
& & Log-cosh & \expstat{\textbf{84.67}}{2.49} & \expstat{\textbf{100.00}}{0.00} & \expstat{\textbf{64.67}}{8.38} & \expstat{\textbf{98.67}}{1.89} & \expstat{\textbf{80.67}}{0.94} & \expstat{\textbf{99.33}}{0.94} & \expstat{\textbf{70.67}}{0.94} & \expstat{\textbf{91.33}}{2.49} \\
\bottomrule
\end{tabular}%
}
\vspace{-2mm}
\end{table*}

We evaluate \textbf{ReJEPA} across three latent world model families spanning simulated navigation, manipulation, continuous control, and real-world autonomous driving. Our experiments test three core hypotheses: (1) bounding large prediction residuals while maintaining smooth origin curvature improves downstream planning over unbounded quadratic (MSE) losses ($u^2$, $\tfrac12 u^2$) and non-smooth or piecewise alternatives ($L_1$, Smooth-$L_1$); (2) the gains of $\log\cosh$ scale with environmental complexity and observation noise, culminating in real-world autonomous driving; and (3) relaxing unconstrained error minimization on outlier transitions allows the encoder to better capture physical states and align with the true planning cost landscape.

\subsection{Experimental Setup}
\label{sec:exp:setup}

\noindent\textbf{Benchmarks and Model Families.}
We first evaluate goal-conditioned planning in simulated control tasks following Temporal Straightening (TS)~\citep{wang2026ts} and LeWorldModel (LeWM)~\citep{maes2026leworldmodel}. In TS (one-step prediction on stopped targets with temporal-curvature regularization), we evaluate Wall, PointMaze--UMaze, PointMaze--Medium, and PushT using either a frozen DINOv2 backbone with a trainable projector or a ResNet trained from scratch, across global ($1\times384$) and spatial ($14\times14\times8$) latent shapes. In LeWM (end-to-end encoder and causal predictor with SIGReg distributional regularization), we evaluate TwoRoom (a two-room layout analogous to Wall), PushT, Reacher, and Cube. We then scale to real-world multi-frame driving on NAVSIM~\citep{dauner2024navsim} using two V-JEPA~2.1~\citep{mur2026v} architectures: a convolutional HENet-TinyM~\citep{xia2024henet} model evaluated on perception MAE and open-loop trajectory errors (ADE/FDE), and a ViT-L/16 backbone paired with a perception-based planner evaluated on the closed-loop Predictive Driver Model Score (PDMS).

\noindent\textbf{Controlled Comparison Protocol.}
Across simulated benchmarks, we compare five training-time predictive penalties: standard squared error ($u^2$), the locally curvature-matched quadratic control ($\tfrac12 u^2$), absolute error ($L_1$, $|u|$), Smooth-$L_1$ (unit-threshold Huber loss, $\beta=1$), and $\log\cosh(u)$. Within each family, network architectures, regularizer weights $\lambda$, training schedules, and inference-time planners (gradient descent for TS, CEM for LeWM, both minimizing squared Euclidean latent distance to the goal) remain strictly fixed. We report open-loop and MPC success rates over three independent world-model runs (three data-sampling seeds in TS and three random initialization seeds in LeWM, each training a separate model) with 50 test episodes per seed. Full protocols are in Appendix~\ref{app:experimental-details}.

\begin{table*}[h]
\centering
\caption{LeWM planning success rates (\%) using CEM across navigation, manipulation, and continuous control tasks. Values are mean $\pm$ standard deviation over three training seeds (50 episodes per seed). Best means are in \textbf{bold}.}
\vspace{-2mm}
\label{tab:leworldmodel_results}
\setlength{\tabcolsep}{6pt}
\renewcommand{\arraystretch}{0.9}
\resizebox{0.85\textwidth}{!}{%
\begin{tabular}{l|cccccccc}
\toprule
\multirow{2}{*}{Predictive penalty}
& \multicolumn{2}{c}{TwoRoom} & \multicolumn{2}{c}{PushT}
& \multicolumn{2}{c}{Reacher} & \multicolumn{2}{c}{Cube} \\
\cmidrule(lr){2-3}\cmidrule(lr){4-5}\cmidrule(lr){6-7}\cmidrule(lr){8-9}
& Open-loop & MPC & Open-loop & MPC & Open-loop & MPC & Open-loop & MPC \\
\midrule
$u^2$ & \expstat{90.67}{2.49} & \expstat{96.67}{1.89} & \expstat{91.33}{2.49} & \expstat{96.67}{2.49} & \expstat{80.67}{4.11} & \expstat{90.67}{3.40} & \expstat{70.67}{4.11} & \expstat{82.67}{4.11} \\
$\frac12u^2$ & \expstat{91.33}{1.89} & \expstat{97.33}{1.89} & \expstat{90.67}{3.40} & \expstat{95.33}{2.49} & \expstat{82.67}{3.40} & \expstat{92.67}{2.49} & \expstat{72.67}{3.77} & \expstat{84.67}{3.40} \\
$L_1$ ($|u|$) & \expstat{92.67}{2.49} & \expstat{96.67}{1.89} & \expstat{92.67}{2.49} & \expstat{96.00}{1.63} & \expstat{84.00}{3.27} & \expstat{94.00}{1.63} & \expstat{72.00}{2.83} & \expstat{87.33}{3.40} \\
Smooth-$L_1$ & \expstat{92.00}{1.63} & \expstat{98.00}{1.63} & \expstat{92.00}{3.27} & \expstat{97.33}{1.89} & \expstat{85.33}{2.49} & \expstat{93.33}{2.49} & \expstat{73.33}{2.49} & \expstat{86.00}{3.27} \\
Log-cosh & \expstat{\textbf{94.67}}{1.89} & \expstat{\textbf{98.67}}{0.94} & \expstat{\textbf{93.33}}{2.49} & \expstat{\textbf{98.67}}{0.94} & \expstat{\textbf{86.67}}{3.40} & \expstat{\textbf{95.33}}{1.89} & \expstat{\textbf{74.67}}{3.40} & \expstat{\textbf{88.67}}{2.49} \\
\bottomrule
\end{tabular}%
}
\vspace{-2mm}
\end{table*}

\subsection{Simulated Control Across Complexity Scales}
\label{sec:exp:planning}

\noindent\textbf{Temporal Straightening.}
\Cref{tab:main_results} shows that $\log\cosh$ outperforms $u^2$ and $\tfrac12 u^2$ across all encoders, latent shapes, and execution modes, while consistently surpassing $L_1$ and Smooth-$L_1$. Notably, the gains grow with task complexity. On the clean Wall navigation task, where dynamics are simple and prediction residuals are small, open-loop gains over $u^2$ are $4.00$ to $5.33$ percentage points. On multi-corridor navigation (U-Maze, Medium) and contact-rich manipulation (PushT), where sharp turns and abrupt block contacts produce large outlier residuals, open-loop success with DINOv2 spatial latents rises from $82.00\%$ to $94.00\%$ on U-Maze and from $72.00\%$ to $84.00\%$ on Medium ($+12.00$ points each; see optimization curves in \Cref{fig:planning_curves}). Averaged across all four tasks (Table~\ref{tab:app-macro} in Appendix~\ref{app:full-results}), $\log\cosh$ improves macro open-loop success by $+5.50$ to $+8.00$ percentage points over $u^2$ and by $+3.17$ to $+6.33$ points over $\tfrac12 u^2$, with spatial latents ($14^2\times8$) exhibiting the largest gains due to token-level attenuation of localized contact errors. Comparing bounded losses confirms why both properties of $\log\cosh$ matter: while $L_1$ prevents outlier frames from dominating updates, its constant unit gradient near zero introduces jitter around well-predicted states and degrades gradient-based planning. Smooth-$L_1$ restores quadratic behavior near zero, but its piecewise second-derivative jump at $|u|=1$ still trails the smooth $C^\infty$ saturation of $\log\cosh$.

\begin{figure}[h]
\centering
\includegraphics[width=\columnwidth]{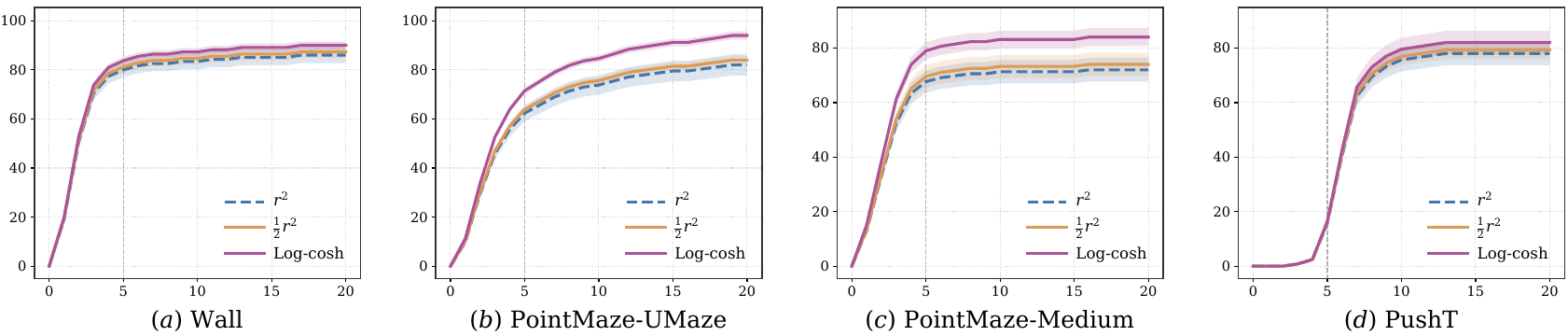}
\vspace{-4mm}
\caption{Open-loop planning success curves with DINOv2 spatial latents ($14\times14\times8$) on Wall, PointMaze--UMaze, PointMaze--Medium, and PushT across gradient-descent optimization iterations.}
\label{fig:planning_curves}
\vspace{-2mm}
\end{figure}

\noindent\textbf{LeWorldModel.}
\Cref{tab:leworldmodel_results} evaluates the penalty replacement in LeWM, where the encoder and multi-step predictor are trained jointly with SIGReg and evaluated with CEM planning. Across all four tasks and both execution modes, $\log\cosh$ achieves the highest success rate. Consistent with TS, improvements are larger on multi-joint control and 3D manipulation than on simple two-room navigation, raising Reacher open-loop success from $80.67\%$ to $86.67\%$ ($+6.00\%$) and Cube MPC success from $82.67\%$ to $88.67\%$ ($+6.00\%$). 


\begin{table}[h]
\centering
\caption{Longer-horizon planning success rates (\%). Values are mean $\pm$ population std over three seeds (50 episodes per seed). Best means are in \textbf{bold}.}
\vspace{-2mm}
\label{tab:long_horizon}
\setlength{\tabcolsep}{3.5pt}
\renewcommand{\arraystretch}{0.9}
\resizebox{\columnwidth}{!}{%
\begin{tabular}{llcccc}
\toprule
\multirow{2}{*}{Encoder}
& \multirow{2}{*}{Penalty}
& \multicolumn{2}{c}{PushT}
& \multicolumn{2}{c}{PointMaze--Medium} \\
\cmidrule(lr){3-4}\cmidrule(lr){5-6}
& & Open-loop & MPC & Open-loop & MPC \\
\midrule
\multirow{5}{*}{DINOv2}
& $u^2$ & \expstat{10.67}{2.49} & \expstat{28.67}{6.60} & \expstat{64.67}{4.99} & \expstat{86.67}{4.11} \\
& $\frac{1}{2}u^2$ & \expstat{12.67}{3.40} & \expstat{30.67}{5.73} & \expstat{67.33}{5.73} & \expstat{89.33}{3.40} \\
& $L_1$ ($|u|$) & \expstat{13.33}{2.49} & \expstat{31.33}{5.73} & \expstat{69.33}{5.73} & \expstat{90.00}{3.27} \\
& Smooth-$L_1$ & \expstat{12.00}{3.27} & \expstat{32.67}{5.73} & \expstat{68.00}{4.90} & \expstat{91.33}{2.49} \\
& Log-cosh & \expstat{\textbf{14.67}}{4.11} & \expstat{\textbf{33.33}}{5.73} & \expstat{\textbf{70.67}}{5.73} & \expstat{\textbf{92.67}}{3.40} \\
\addlinespace[2pt]
\multirow{5}{*}{ResNet}
& $u^2$ & \expstat{8.67}{3.40} & \expstat{30.67}{7.72} & \expstat{62.67}{6.60} & \expstat{87.33}{3.77} \\
& $\frac{1}{2}u^2$ & \expstat{10.67}{4.11} & \expstat{33.33}{6.80} & \expstat{65.33}{5.25} & \expstat{90.67}{4.11} \\
& $L_1$ ($|u|$) & \expstat{10.00}{3.27} & \expstat{35.33}{5.73} & \expstat{66.00}{4.90} & \expstat{92.00}{3.27} \\
& Smooth-$L_1$ & \expstat{11.33}{4.11} & \expstat{34.00}{6.53} & \expstat{67.33}{5.73} & \expstat{91.33}{3.40} \\
& Log-cosh & \expstat{\textbf{12.67}}{5.73} & \expstat{\textbf{36.67}}{6.60} & \expstat{\textbf{68.67}}{6.80} & \expstat{\textbf{93.33}}{2.49} \\
\bottomrule
\end{tabular}%
}
\vspace{-3mm}
\end{table}

\noindent\textbf{Longer-Horizon Planning.}
\Cref{tab:long_horizon} tests whether the gains hold over extended planning horizons on PushT and PointMaze--Medium, where multi-step rollouts accumulate greater prediction uncertainty. Across both encoders and execution modes, $\log\cosh$ maintains a $4.00$ to $6.00$ percentage point lead over $u^2$ and consistently outperforms $\tfrac12 u^2$, $L_1$, and Smooth-$L_1$, raising ResNet MPC success from $30.67\%$ to $36.67\%$ on PushT and to $93.33\%$ on Medium.



\subsection{Scaling to Real-World Autonomous Driving on NAVSIM}
\label{sec:exp:driving}

Real-world driving videos contain severe background clutter, dynamic traffic occlusions, and multi-camera viewpoint changes, making unconstrained error minimization across all spatiotemporal tokens particularly harmful. We evaluate our method on NAVSIM~\citep{dauner2024navsim} across two architectures and downstream protocols (full details in Appendix~\ref{app:driving-details}).

\begin{table}[h]
\centering
\caption{Multi-task driving perception MAE and open-loop planning errors (ADE/FDE) with HENet-TinyM on NAVSIM, with and without pretraining. Lower is better.}
\vspace{-2mm}
\label{tab:driving_results}
\setlength{\tabcolsep}{3.2pt}
\renewcommand{\arraystretch}{0.9}
\resizebox{\columnwidth}{!}{%
\begin{tabular}{clrrrrrrr}
\toprule
\multirow{2}{*}{Pretrained} & \multirow{2}{*}{Loss}
& \multicolumn{5}{c}{Perception: MAE $\downarrow$}
& \multicolumn{2}{c}{Planning $\downarrow$} \\
\cmidrule(lr){3-7}\cmidrule(lr){8-9}
 & & Veh. & Ped. & Bic. & Cone & Barrier & ADE/m & FDE/m \\
\midrule
\multirow{3}{*}{\cmark}
& $u^2$ & 2.93 & 3.79 & 0.053 & 1.02 & 0.158 & 7.91 & 14.21 \\
& $\tfrac12 u^2$ & 2.94 & 3.81 & 0.053 & 1.02 & 0.158 & 7.95 & 14.28 \\
& Log-cosh & \textbf{2.65} & \textbf{3.43} & \textbf{0.048} & \textbf{0.91} & \textbf{0.133} & \textbf{7.15} & \textbf{12.85} \\
\addlinespace[2pt]
\multirow{3}{*}{\xmark}
& $u^2$ & 3.08 & 3.98 & 0.055 & 1.07 & 0.165 & 8.36 & 15.16 \\
& $\tfrac12 u^2$ & 3.09 & 4.00 & 0.055 & 1.07 & 0.166 & 8.40 & 15.24 \\
& Log-cosh & \textbf{2.71} & \textbf{3.50} & \textbf{0.049} & \textbf{0.93} & \textbf{0.136} & \textbf{7.30} & \textbf{13.12} \\
\bottomrule
\end{tabular}%
}
\vspace{-3mm}
\end{table}

\begin{table*}[h]
\centering
\caption{Closed-loop driving evaluation on NAVSIM \texttt{navtest} (12,146 scenes) using a V-JEPA~2.1 ViT-L/16 backbone and perception-based planner. Sub-scores are No at-fault Collisions (NC), Drivable Area Compliance (DAC), Driving Direction Compliance (DDC), Time-to-Collision (TTC), Comfort (Comf.), and Ego Progress (EP).}
\vspace{-2mm}
\label{tab:driving_pdms}
\setlength{\tabcolsep}{6pt}
\renewcommand{\arraystretch}{0.9}
\resizebox{0.94\textwidth}{!}{%
\begin{tabular}{ll|cccccc|c}
\toprule
Checkpoint & Pretraining Loss & NC $\uparrow$ & DAC $\uparrow$ & DDC $\uparrow$ & TTC $\uparrow$ & Comf. $\uparrow$ & EP $\uparrow$ & \textbf{PDMS} $\uparrow$ \\
\midrule
Best Val (Epoch 14) & Matched MSE ($\tfrac12 u^2$) & 98.43 & 96.95 & \textbf{95.54} & 94.90 & 99.78 & 86.96 & 90.48 \\
Best Val (Epoch 13) & \textbf{Log-cosh} & \textbf{98.53} & \textbf{97.46} & 95.14 & \textbf{95.06} & \textbf{99.86} & \textbf{87.19} & \textbf{90.91} \\
\midrule
Epoch 0 (331 steps) & $\tfrac12 u^2 \rightarrow$ \textbf{Log-cosh} & $96.5 \rightarrow \mathbf{97.0}$ & $89.5 \rightarrow \mathbf{95.2}$ & $\mathbf{97.8} \rightarrow 97.7$ & $90.4 \rightarrow \mathbf{91.6}$ & $76.9 \rightarrow \mathbf{77.2}$ & $58.2 \rightarrow \mathbf{65.2}$ & $69.03 \rightarrow \mathbf{75.35}$ (\textbf{+6.32}) \\
Epoch 3 (1,324 steps) & $\tfrac12 u^2 \rightarrow$ \textbf{Log-cosh} & $\mathbf{98.6} \rightarrow 98.1$ & $93.6 \rightarrow \mathbf{94.5}$ & $\mathbf{96.9} \rightarrow 96.5$ & $93.1 \rightarrow \mathbf{93.7}$ & $\mathbf{98.6} \rightarrow 97.5$ & $77.9 \rightarrow \mathbf{80.7}$ & $83.87 \rightarrow \mathbf{85.58}$ (\textbf{+1.71}) \\
Epoch 5 (1,986 steps) & $\tfrac12 u^2 \rightarrow$ \textbf{Log-cosh} & $\mathbf{98.7} \rightarrow 98.6$ & $\mathbf{96.7} \rightarrow 96.6$ & $\mathbf{96.8} \rightarrow 96.5$ & $\mathbf{95.6} \rightarrow 95.3$ & $\mathbf{99.8} \rightarrow 98.2$ & $82.2 \rightarrow \mathbf{83.5}$ & $88.66 \rightarrow \mathbf{88.74}$ (\textbf{+0.08}) \\
Epoch 7 (2,648 steps) & $\tfrac12 u^2 \rightarrow$ \textbf{Log-cosh} & $98.6 \rightarrow \mathbf{98.7}$ & $94.8 \rightarrow \mathbf{97.2}$ & $\mathbf{96.5} \rightarrow 96.1$ & $94.9 \rightarrow \mathbf{95.9}$ & $\mathbf{99.9} \rightarrow 99.8$ & $82.0 \rightarrow \mathbf{83.9}$ & $87.21 \rightarrow \mathbf{89.77}$ (\textbf{+2.56}) \\
\bottomrule
\end{tabular}%
}
\vspace{-2mm}
\end{table*}

\begin{figure}[h]
\centering
\vspace{-0.3cm}
\includegraphics[width=\columnwidth]{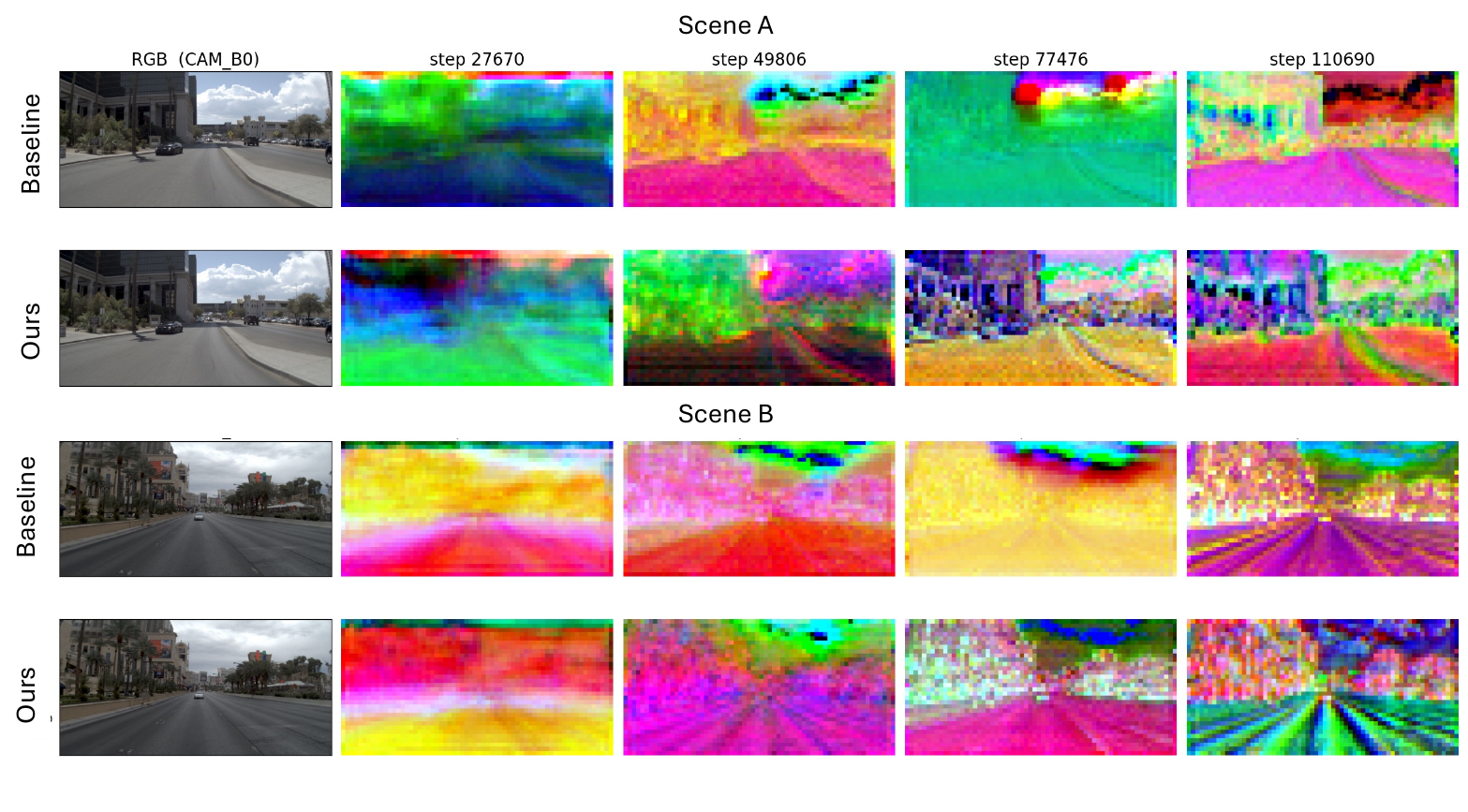}
\vspace{-0.8cm}
\caption{Latent feature visualizations on NAVSIM driving scenes. Baseline and Ours denote $u^2$ and $\log\cosh$.}
\label{fig:feature_visualization}
\vspace{-0.2cm}
\end{figure}

\noindent\textbf{Multi-Task Perception and Open-Loop Trajectory Errors.}
We first train a convolutional JEPA model (HENet-TinyM~\citep{xia2024henet} context encoder, EMA target encoder, and 3D predictor) on 708,416 eight-frame ($3.5\,\mathrm{s}$ at $2\,\mathrm{Hz}$) single-camera clips from NAVSIM \texttt{trainval} across eight views under tube masking ($43\%$--$49\%$ masked tokens). The training objective combines masked feature prediction with boundary-distance-weighted visible context reconstruction after affine-free channel-wise LayerNorm. As shown in \Cref{tab:driving_results}, $\tfrac12 u^2$ performs nearly identically to $u^2$ because uniform scaling cancels in AdamW. In contrast, $\log\cosh$ achieves the lowest error across all seven metrics in both pretraining regimes, lowering perception MAEs across all five object categories by $9.4\%$--$17.6\%$ and reducing open-loop planning ADE/FDE from $7.91/14.21\,\mathrm{m}$ to $7.15/12.85\,\mathrm{m}$ (pretrained) and from $8.36/15.16\,\mathrm{m}$ to $7.30/13.12\,\mathrm{m}$ (from scratch). Feature maps in \Cref{fig:feature_visualization} show cleaner structural boundaries around road elements under $\log\cosh$.

\noindent\textbf{End-to-End Closed-Loop Planning with V-JEPA~2.1 ViT-L.}
We next scale pretraining on NAVSIM \texttt{trainval} to a V-JEPA~2.1 ViT-L/16 encoder (with RoPE, a 24-layer predictor, and 4-layer hierarchical target distillation under 3D multi-block masking of 8 small and 2 large blocks; global batch size 640). Here, the baseline MSE penalty is natively formulated as the curvature-matched quadratic $\tfrac12\|z - h\|_2^2$, isolating large-error saturation as the sole difference when switching to $\log\cosh$. For downstream evaluation, we attach a Drive-JEPA perception-based planning head to the EMA target encoder. Taking two consecutive front-camera frames and ego-vehicle status (command, velocity, acceleration), the planner initializes $32\times 8$ learnable queries representing 32 trajectory proposals over a $4.0\,\mathrm{s}$ horizon ($0.5\,\mathrm{s}$ intervals) and refines them through four shared BEVFormer-style deformable self-attention and spatial cross-attention layers, followed by an MLP scorer supervised online by the PDM simulator. We fine-tune the full model end-to-end on \texttt{navtrain} (84,792 training and 18,136 validation samples), select the best validation checkpoint on \texttt{navval}, and evaluate on all 12,146 scenes of the official \texttt{navtest} split using closed-loop PDMS (\Cref{tab:driving_pdms}).

\begin{figure}[h]
\centering
\includegraphics[width=0.75\columnwidth]{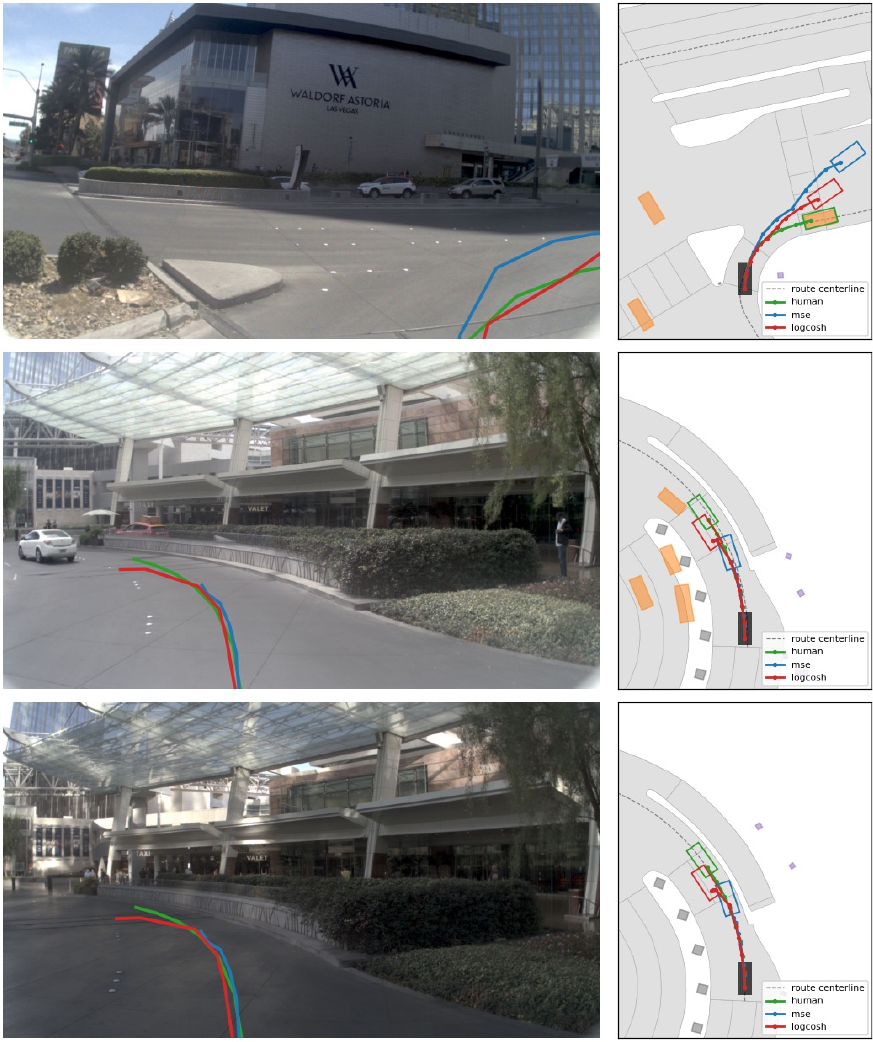}
\vspace{-1mm}
\caption{Qualitative closed-loop planning comparison on a curved urban roadway in NAVSIM \texttt{navtest} across camera views and bird's-eye-view (BEV) maps. We compare the human trajectory (green) against plans from V-JEPA~2.1 ViT-L encoders pretrained with matched MSE (blue) and $\log\cosh$ (red). While MSE drifts toward the inner curb and terminates early, $\log\cosh$ smoothly follows the route centerline (dashed gray) and closely matches the human reference.}
\label{fig:navsim_traj_vis}
\vspace{-3mm}
\end{figure}

\Cref{tab:driving_pdms} reveals three findings: (1) At Epoch~0 (after just 331 fine-tuning steps, when planning performance primarily reflects the pretrained encoder), $\log\cosh$ outperforms MSE by $+6.32$ PDMS ($75.35\%$ vs.\ $69.03\%$), driven by large gains in Drivable Area Compliance ($+5.7\%$) and Ego Progress ($+7.0\%$). By not overfitting transient background clutter during pretraining, $\log\cosh$ yields a representation where road geometry and ego corridors are immediately accessible. (2) Whereas the MSE model drops by $1.45$ points at Epoch~7 due to DAC degradation, $\log\cosh$ improves monotonically across all checkpoints. (3) At the best validation checkpoint, $\log\cosh$ reaches $90.91\%$ PDMS vs.\ $90.48\%$ for MSE, leading on five of six sub-metrics.

\noindent\textbf{Qualitative Trajectory Comparison.}
\Cref{fig:navsim_traj_vis} illustrates closed-loop planning behavior on a curved roadway in \texttt{navtest} featuring strong architectural reflections, shadows, and vegetation clutter. Under this visual complexity, the encoder pretrained with matched MSE (blue) veers prematurely toward the right curb boundary and exhibits conservative forward progress. In contrast, the $\log\cosh$-pretrained encoder (red) tracks the curved route centerline smoothly across both camera projections and bird's-eye-view (BEV) coordinates, aligning closely with the human driver trajectory (green). This qualitative behavior directly mirrors the quantitative gains in Drivable Area Compliance and Ego Progress reported in \Cref{tab:driving_pdms}.


\begin{table*}[h]
\centering
\caption{Physical-state probing on PushT with frozen LeWM encoders (mean $\pm$ std across three probe seeds).}
\vspace{-2mm}
\label{tab:probe-pusht}
\setlength{\tabcolsep}{5pt}
\renewcommand{\arraystretch}{0.9}
\resizebox{0.95\textwidth}{!}{%
\begin{tabular}{
l@{\hspace{8pt}}|@{\hspace{8pt}}
cc@{\hspace{6pt}}cc@{\hspace{10pt}}
cc@{\hspace{6pt}}cc@{\hspace{10pt}}
cc@{\hspace{6pt}}cc
}
\toprule
\multirow{2}{*}{Method}
& \multicolumn{4}{c}{Agent Location (Linear / MLP)}
& \multicolumn{4}{c}{Block Location (Linear / MLP)}
& \multicolumn{4}{c}{Block Angle (Linear / MLP)} \\
\cmidrule(lr){2-5}\cmidrule(lr){6-9}\cmidrule(lr){10-13}
& MSE $\downarrow$ & $r\uparrow$ & MSE $\downarrow$ & $r\uparrow$
& MSE $\downarrow$ & $r\uparrow$ & MSE $\downarrow$ & $r\uparrow$
& MSE $\downarrow$ & $r\uparrow$ & MSE $\downarrow$ & $r\uparrow$ \\
\midrule
LeWM with $u^2$
& 0.0513 $\pm$ {\tiny 0.0051} & 0.9743 & 0.0041 $\pm$ {\tiny 0.0005} & 0.9979
& 0.0287 $\pm$ {\tiny 0.0035} & 0.9857 & 0.0011 $\pm$ {\tiny 0.0002} & 0.9991
& 0.1823 $\pm$ {\tiny 0.0173} & 0.9053 & 0.0207 $\pm$ {\tiny 0.0021} & 0.9897 \\
LeWM with $\frac{1}{2}u^2$
& 0.0497 $\pm$ {\tiny 0.0047} & 0.9751 & 0.0038 $\pm$ {\tiny 0.0004} & 0.9981
& 0.0263 $\pm$ {\tiny 0.0033} & 0.9871 & 0.0010 $\pm$ {\tiny 0.0002} & 0.9992
& 0.1733 $\pm$ {\tiny 0.0167} & 0.9097 & 0.0193 $\pm$ {\tiny 0.0023} & 0.9907 \\
\textbf{LeWM with Log-cosh}
& \textbf{0.0383} $\pm$ {\tiny 0.0039} & \textbf{0.9817} & \textbf{0.0024} $\pm$ {\tiny 0.0003} & \textbf{0.9993}
& \textbf{0.0203} $\pm$ {\tiny 0.0027} & \textbf{0.9903} & \textbf{0.0007} $\pm$ {\tiny 0.0001} & \textbf{0.9995}
& \textbf{0.1373} $\pm$ {\tiny 0.0137} & \textbf{0.9277} & \textbf{0.0143} $\pm$ {\tiny 0.0017} & \textbf{0.9933} \\
\bottomrule
\end{tabular}%
}
\vspace{-2mm}
\end{table*}

\subsection{Physical Structure of the Latent Space}
\label{sec:exp:physical_structure}


\begin{figure}[h]
\centering
\includegraphics[width=\columnwidth]{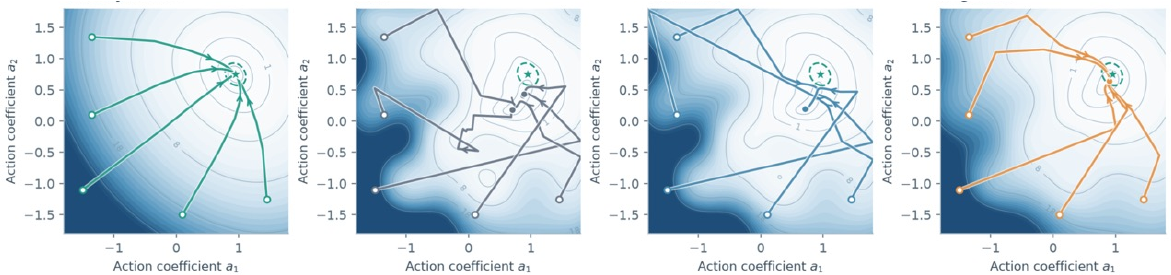}
\vspace{-0.7cm}
\caption{Action-space optimization trajectories under true dynamics, $u^2$, $\tfrac12 u^2$, and $\log\cosh$ (left to right) across the $(a_1, a_2)$ plane.}
\label{fig:action_optimization}
\vspace{-3mm}
\end{figure}

\begin{figure}[h]
\centering
\includegraphics[width=\columnwidth]{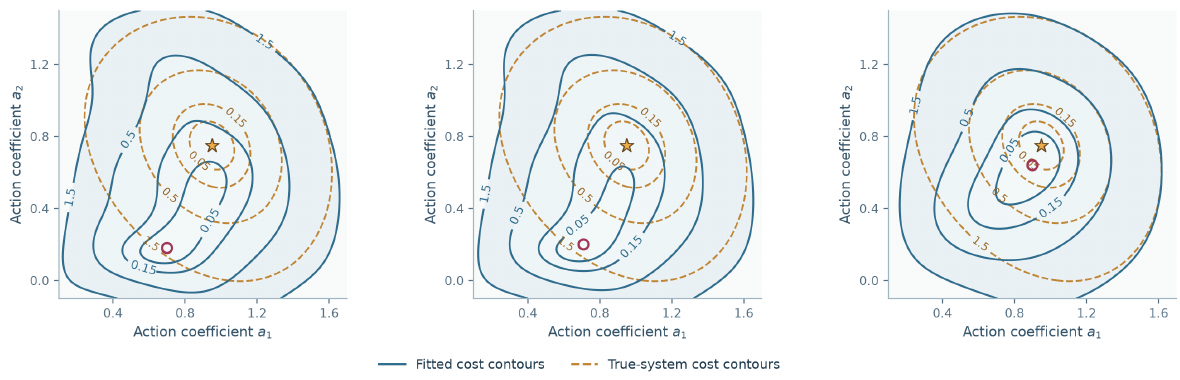}
\vspace{-0.7cm}
\caption{Learned planning cost contours (solid blue) vs.\ system reference contours (dashed orange) for $u^2$, $\tfrac12 u^2$, and $\log\cosh$.}
\label{fig:action_cost_contours}
\vspace{-3mm}
\end{figure}

Downstream planning success reflects the combined performance of the encoder, the predictor, and the action-search procedure. To directly test whether bounding outlier errors preserves more physical state information in the encoder itself, we follow the LeWM probing protocol~\citep{maes2026leworldmodel} on PushT (\Cref{tab:probe-pusht}; details in Appendix~\ref{app:probe-details}). We freeze the trained LeWM encoders and train linear and MLP probes on a fixed split to recover ground-truth agent coordinates, block coordinates (scaled to $[-1, 1]$), and block orientation $\theta$ (encoded continuously as $(\cos\theta, \sin\theta)$). Relative to $u^2$ and $\tfrac12 u^2$, $\log\cosh$ reduces linear-probe MSE by $22.9\%$--$25.3\%$ for agent location, $22.8\%$--$29.3\%$ for block location, and $20.8\%$--$24.7\%$ for block orientation. With MLP probes, relative MSE reductions reach $36.8\%$--$41.5\%$ for agent location and $25.9\%$--$36.4\%$ for block pose while achieving the highest Pearson correlations ($r \ge 0.9933$). Relaxing the penalty on noisy contact transitions frees encoder capacity to represent core system states across readout capacities.


\subsection{Latent Distance and Action-Space Planning Geometry}
\label{sec:exp:visualization}
\label{sec:exp:action_geometry}

\begin{figure}[h]
\centering
\includegraphics[width=0.7\columnwidth]{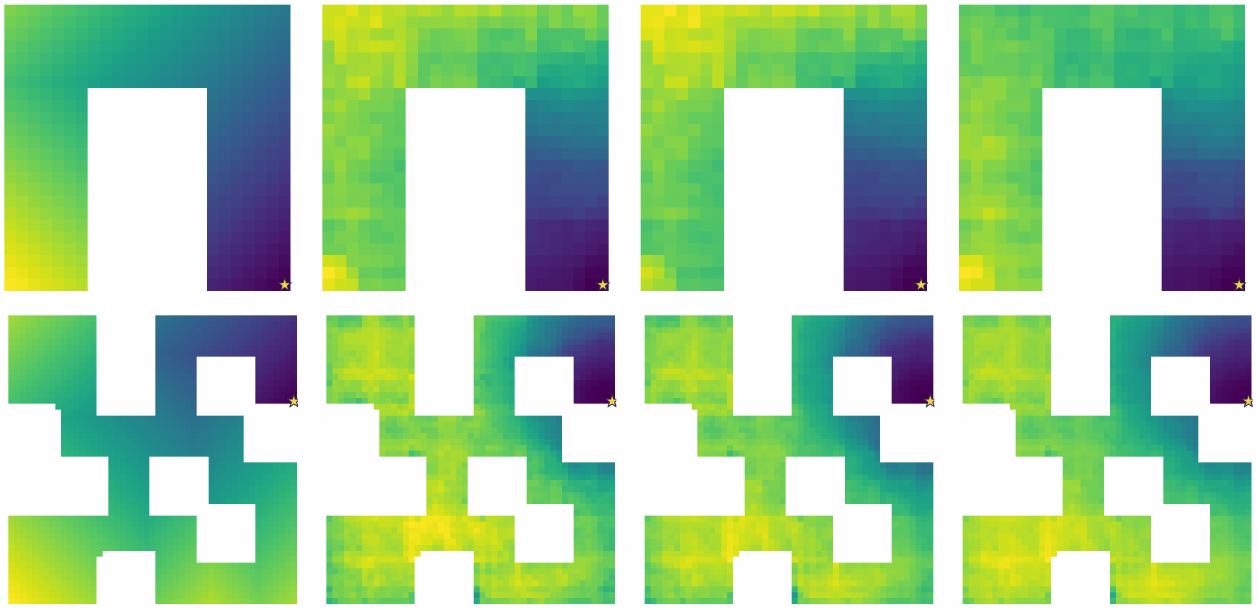}
\vspace{-0.2cm}
\caption{Latent goal-distance heatmaps for PointMaze--UMaze (top four) and PointMaze--Medium (bottom four): ground truth, $u^2$, $\tfrac{1}{2}u^2$, and $\log\cosh$ (left to right). White denotes walls; the yellow star marks the goal.}
\label{fig:latent_distance}
\vspace{-4mm}
\end{figure}

\noindent\textbf{Latent Goal-Distance Heatmaps.}
Because a learned probe can recover state variables even when raw Euclidean latent distances guide planning poorly, \Cref{fig:latent_distance} directly inspects the planner's native squared latent distance $\|f_\theta(o) - f_\theta(o_{\mathrm{goal}})\|_2^2$ across navigable states in PointMaze--UMaze and PointMaze--Medium, where states separated by a thin wall are visually similar in global coordinates but dynamically distant. Under $u^2$ and $\tfrac12 u^2$, distance contours bleed across maze walls, creating false low-cost regions on the opposite side of obstacles. Under $\log\cosh$, latent distances respect corridor topology and decay smoothly along valid paths toward the goal.

\noindent\textbf{Action-Space Trajectories and Cost Contours.}
Finally, \Cref{fig:action_optimization,fig:action_cost_contours} examine the composed multi-step planning objective $\mathcal J(a) = \|\hat Z(a) - Z_{\mathrm{goal}}\|_2^2$ (combining the encoder, predictor rollouts, and goal cost) across a 2D action subspace $(a_1, a_2)$. In \Cref{fig:action_optimization}, gradient optimization trajectories under $u^2$ and $\tfrac12 u^2$ stall or diverge along spurious local ridges, whereas $\log\cosh$ forms a well-conditioned basin that guides trajectories directly to the optimal action. In \Cref{fig:action_cost_contours}, the learned cost contours of $\log\cosh$ match both the center and elliptical orientation of the true-system reference far more closely than quadratic baselines.
\section{Conclusion}
\label{sec:conclusion}

In latent world models, forcing every multi-frame prediction residual to zero under unbounded $L_2$ loss allows noisy transitions to distort the representation, whereas non-smooth $L_1$ penalties disrupt the local curvature required for planning. By combining bounded large-error gradients with smooth quadratic behavior near zero via the hyperparameter-free $\log\cosh$ loss, \textbf{ReJEPA} protects core physical structure and reduces gradient conflict with structural regularizers without extra parameters or inference cost. Across Temporal Straightening, LeWorldModel, and V-JEPA~2.1, this training-time replacement consistently outperforms quadratic, $L_1$, and Smooth-$L_1$ baselines, with gains scaling from simulated control to real-world driving on NAVSIM. Probing and geometric analyses further confirm improved physical state accessibility and closer alignment with true planning cost landscapes, suggesting bounded predictive regulation as a simple, general principle for scaling latent world models.

\clearpage
\newpage
{

\subsection*{LLM Usage Statement}

We used OpenAI Codex for language polishing of the manuscript and assistance with code debugging. The authors take full responsibility for all submitted content.

    \small
    \bibliographystyle{ieeenat_fullname}
    \bibliography{main}
}

\clearpage
\section{Analytical Details}
\label{app:analytical-details}

\paragraph{Numerically stable evaluation and derivatives.}
Directly computing $\cosh(u) = \tfrac12(e^u + e^{-u})$ in floating-point arithmetic can overflow when $|u|$ is large. For all $u \in \mathbb R$, an algebraically equivalent and numerically stable identity is
\begin{equation}
\log\cosh(u)
=|u|+\operatorname{softplus}(-2|u|)-\log 2,
\label{eq:app-analytic-stable-logcosh}
\end{equation}
where $\operatorname{softplus}(x)=\log(1+e^x)$. To verify Equation~\ref{eq:app-analytic-stable-logcosh}, note that $\cosh(u) = \cosh(|u|) = \tfrac12 e^{|u|}(1 + e^{-2|u|})$. Taking the natural logarithm gives
\[
\begin{aligned}
\log\cosh(u)
&= \log\!\left(\tfrac12 e^{|u|}(1 + e^{-2|u|})\right) \\
&= |u| + \log(1 + e^{-2|u|}) - \log 2 \\
&= |u| + \operatorname{softplus}(-2|u|) - \log 2.
\end{aligned}
\]
Because $-2|u| \le 0$ for all $u \in \mathbb R$, the exponential $e^{-2|u|}$ lies in $(0, 1]$, avoiding overflow. For $u \neq 0$, differentiating Equation~\ref{eq:app-analytic-stable-logcosh} yields
\[
\begin{aligned}
\psi_{\mathrm{lc}}(u)
&= \frac{\mathrm d}{\mathrm d u}\log\cosh(u)
= \operatorname{sign}(u)\!\left(1 - \frac{2e^{-2|u|}}{1 + e^{-2|u|}}\right) \\
&= \operatorname{sign}(u)\frac{1 - e^{-2|u|}}{1 + e^{-2|u|}}
= \tanh(u),
\end{aligned}
\]
which extends smoothly through $u = 0$ with $\psi_{\mathrm{lc}}(0) = 0$ and second derivative $\rho_{\mathrm{lc}}''(u) = \operatorname{sech}^2(u) = 1 - \tanh^2(u) \in (0, 1]$.

\paragraph{Asymptotic scaling and scalar weights.}
Relative to the curvature-matched quadratic penalty $\rho_{\mathrm{q}}(u) = \tfrac12 u^2$, $\log\cosh$ scales each residual score by $w_{\mathrm{lc}}(u) = \tanh(u)/u$ for $u \neq 0$, with $w_{\mathrm{lc}}(0) = 1$. The fractional attenuation is $\delta(u) = 1 - w_{\mathrm{lc}}(u)$. Using the Maclaurin series $\tanh(u) = u - \tfrac13 u^3 + \tfrac{2}{15}u^5 + O(u^7)$ for $|u| < \pi/2$, the weight and attenuation expand around zero as
\[
\begin{aligned}
w_{\mathrm{lc}}(u) &= 1 - \tfrac13 u^2 + \tfrac{2}{15}u^4 + O(u^6), \\
\delta(u) &= \tfrac13 u^2 - \tfrac{2}{15}u^4 + O(u^6).
\end{aligned}
\]
Thus, small residuals undergo negligible attenuation ($\delta(u) = O(u^2)$ as $u \to 0$), preserving quadratic behavior near convergence. Conversely, as $|u| \to \infty$, $|\tanh(u)| = 1 - 2e^{-2|u|} + O(e^{-4|u|})$, so the weight decays inversely with residual magnitude, $w_{\mathrm{lc}}(u) = 1/|u| + O(e^{-2|u|}/|u|)$, and $\delta(u) \to 1$. Table~\ref{tab:app-analytic-residual-weights} lists exact values across representative magnitudes.

\begin{table}[h]
\centering
\small
\caption{Analytically computed values of $|\tanh(u)|$, effective weight $w_{\mathrm{lc}}(u) = \tanh(u)/u$, and fractional attenuation $\delta(u) = 1 - w_{\mathrm{lc}}(u)$, rounded to six decimal places. These are exact function evaluations rather than empirical measurements.}
\label{tab:app-analytic-residual-weights}
\setlength{\tabcolsep}{6pt}
\begin{tabular}{rrrr}
\toprule
$|u|$ & $|\tanh(u)|$ & $w_{\mathrm{lc}}(u)$ & $\delta(u)$ \\
\midrule
0   & 0.000000 & 1.000000 & 0.000000 \\
0.5 & 0.462117 & 0.924234 & 0.075766 \\
1   & 0.761594 & 0.761594 & 0.238406 \\
2   & 0.964028 & 0.482014 & 0.517986 \\
3   & 0.995055 & 0.331685 & 0.668315 \\
5   & 0.999909 & 0.199982 & 0.800018 \\
\bottomrule
\end{tabular}
\end{table}

\paragraph{Simultaneous local descent condition.}
Section~\ref{sec:method:interaction} analyzes how the predictive gradient $p_\rho = \nabla_\vartheta\mathcal P_\rho$ affects first-order progress on the regularizer $\mathcal G = \lambda\mathcal R$. A complementary question is when the combined raw-gradient step $v = -(p_\rho + h)$ is a strict first-order descent direction for both $\mathcal P_\rho$ and $\mathcal G$ simultaneously.

\begin{proposition}[Raw-gradient common descent]
\label{prop:app-raw-gradient-common-descent}
At a fixed state $\vartheta$ and minibatch realization, let $p = \nabla_\vartheta\mathcal P_\rho(\vartheta)$ and $h = \nabla_\vartheta\mathcal G(\vartheta)$ be nonzero gradients of locally differentiable objectives on shared parameters $\vartheta$, with other parameters fixed and stopped targets held constant. Define the norm ratio and cosine similarity as
\[
\Gamma=\frac{\|h\|_2}{\|p\|_2}>0,\qquad
C=\frac{\langle p,h\rangle}{\|p\|_2\|h\|_2}\in[-1,1].
\]
The combined direction $v = -(p+h)$ is a strict first-order descent direction for both $\mathcal P_\rho$ and $\mathcal G$ if and only if
\begin{equation}
1+\Gamma C>0\qquad\text{and}\qquad\Gamma+C>0.
\label{eq:app-raw-common-descent}
\end{equation}
\end{proposition}

\begin{proof}
The directional derivatives along $v = -(p+h)$ are
\[
\begin{aligned}
D\mathcal P_\rho[v] &= \langle p, -(p+h)\rangle = -\|p\|_2^2(1+\Gamma C), \\
D\mathcal G[v] &= \langle h, -(p+h)\rangle = -\|p\|_2\|h\|_2(\Gamma+C).
\end{aligned}
\]
Since $\|p\|_2 > 0$ and $\|h\|_2 > 0$, both derivatives are strictly negative if and only if $1 + \Gamma C > 0$ and $\Gamma + C > 0$.
\end{proof}

Whenever the predictive and regularizer gradients cooperate ($C \ge 0$), both conditions in Equation~\ref{eq:app-raw-common-descent} hold for any $\Gamma > 0$. When they conflict ($C < 0$), common descent requires $C > -\min(\Gamma, 1/\Gamma)$. Note that $\Gamma + C > 0$ is algebraically equivalent to $\|h\|_2^2 + A_\rho > 0$ in Equation~\ref{eq:representation-progress}, where $A_\rho = \langle p, h \rangle$. Thus, when attenuating large conflicting residuals increases $A_\rho$ (Proposition~\ref{prop:residual-interaction}), it directly widens the margin for decreasing $\mathcal G$ under a joint raw-gradient step.

\providecommand{\appstat}[2]{#1\,{\scriptsize$\pm$\,#2}}
\providecommand{\tbd}[2]{#1\,{\scriptsize$\pm$\,#2}}

\section{Experimental Protocols}
\label{app:experimental-details}

\subsection{Controlled Loss Replacement}
Across all simulated benchmarks in Temporal Straightening (TS)~\citep{wang2026ts} and LeWorldModel (LeWM)~\citep{maes2026leworldmodel}, we vary only the coordinate-wise training penalty $\rho(u)$ among squared error ($u^2$), the curvature-matched quadratic control ($\tfrac12 u^2$), absolute error ($L_1$, $|u|$), Smooth-$L_1$ (Huber loss with $\beta=1$), and $\log\cosh(u)$. The residual vector $r$ incorporates each model's native pre-loss normalization. Within each family, we retain all masks, loss reductions, architectures, representation regularizers and weights $\lambda$, training epochs, checkpoint rules, and inference-time planners (which minimize the squared Euclidean latent distance to the goal). Comparing $\log\cosh(u)$ against $\tfrac12 u^2$ matches origin curvature ($\rho''(0)=1$) and tests whether constant loss rescaling explains the gains. Real-world driving experiments on NAVSIM~\citep{dauner2024navsim} follow the protocols in Appendix~\ref{app:driving-details}.

\paragraph{Simulated model families.}
TS optimizes one-step latent prediction from observation and action history using a stopped-gradient target encoder, temporal-curvature regularization, and gradient descent (GD) planning. We test a frozen DINOv2 ViT backbone with a trainable projector and a ResNet trained from scratch, each with global ($1\times384$) and spatial ($14\times14\times8$) latents. LeWM trains an encoder and a teacher-forced causal predictor end-to-end with SIGReg distributional regularization and plans via the Cross-Entropy Method (CEM). Note that the two-room navigation task is named Wall in TS and TwoRoom in LeWM; we retain each benchmark's native naming.

\paragraph{Training and evaluation protocols.}
TS models train for 20 epochs, and LeWM models use the standard 10-epoch checkpoint. Each configuration is evaluated over three independent world-model runs with 50 test episodes per run. In TS, each run uses an independent data-sampling seed that generates a distinct dataset and trains a separate world model from scratch. In LeWM, each run uses an independent random training seed. Open-loop execution follows the initial plan without feedback, whereas MPC replans iteratively from updated observations.

\subsection{Reporting and Summary Metrics}
\label{app:reporting}
Simulated planning tables report mean and standard deviation across the three world-model runs (with longer-horizon tables reporting population standard deviation). In PushT physical probing, variability ($\pm$) is measured across three probe initialization seeds on frozen encoders.

For success rates (\%), the gain in percentage points (pp) over baseline $b$ is $S_{\mathrm{lc}} - S_b$. For error metrics $E$, the relative reduction is $100 \times (E_b - E_{\mathrm{lc}})/E_b$. Task-macro success is the unweighted mean across the four tasks within a configuration and execution mode.

\section{Complete Simulated Planning Results}
\label{app:full-results}

\begin{figure*}[!t]
\centering
\includegraphics[width=0.92\textwidth]{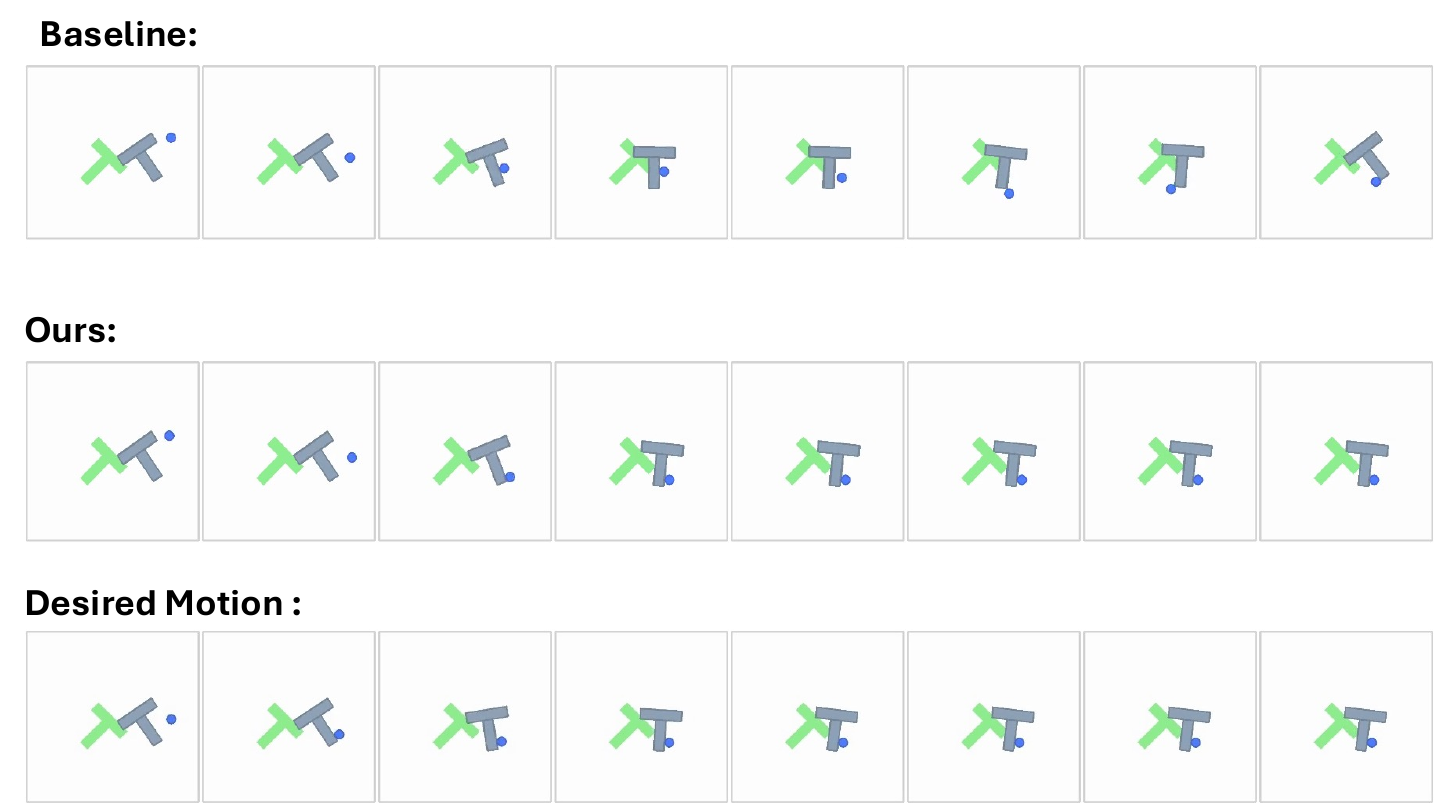}
\vspace{-3mm}
\caption{Qualitative comparison of long-horizon open-loop gradient-descent planning rollouts on PushT.}
\label{fig:app_PushT_visualization}
\vspace{2mm}
\includegraphics[width=0.92\textwidth]{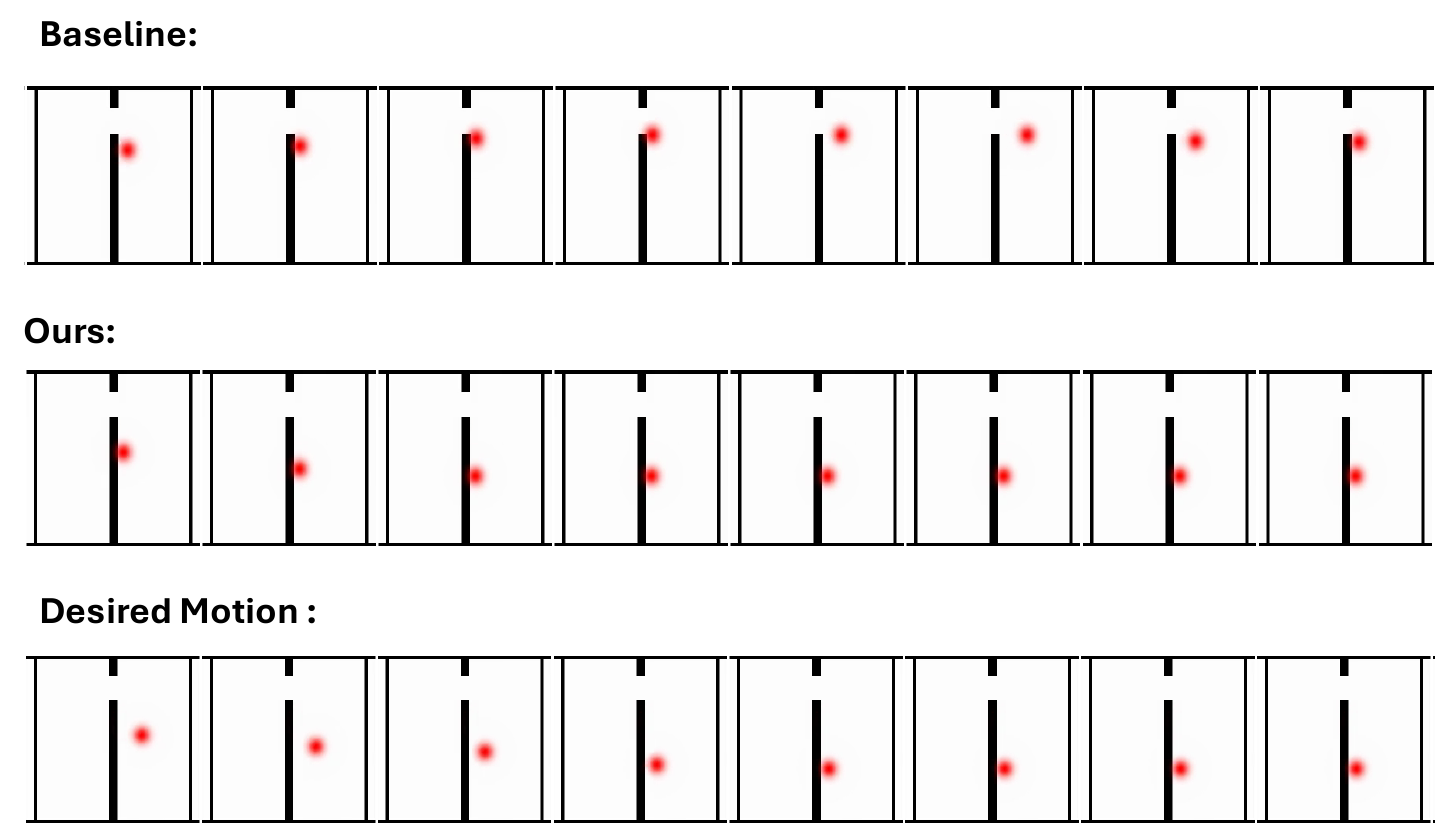}
\vspace{-3mm}
\caption{Open-loop planned trajectories on the Wall (TwoRoom) navigation benchmark.}
\label{fig:app_Tworoom_visualization}
\vspace{-2mm}
\end{figure*}

\begin{figure*}[!t]
\centering
\includegraphics[width=0.92\textwidth]{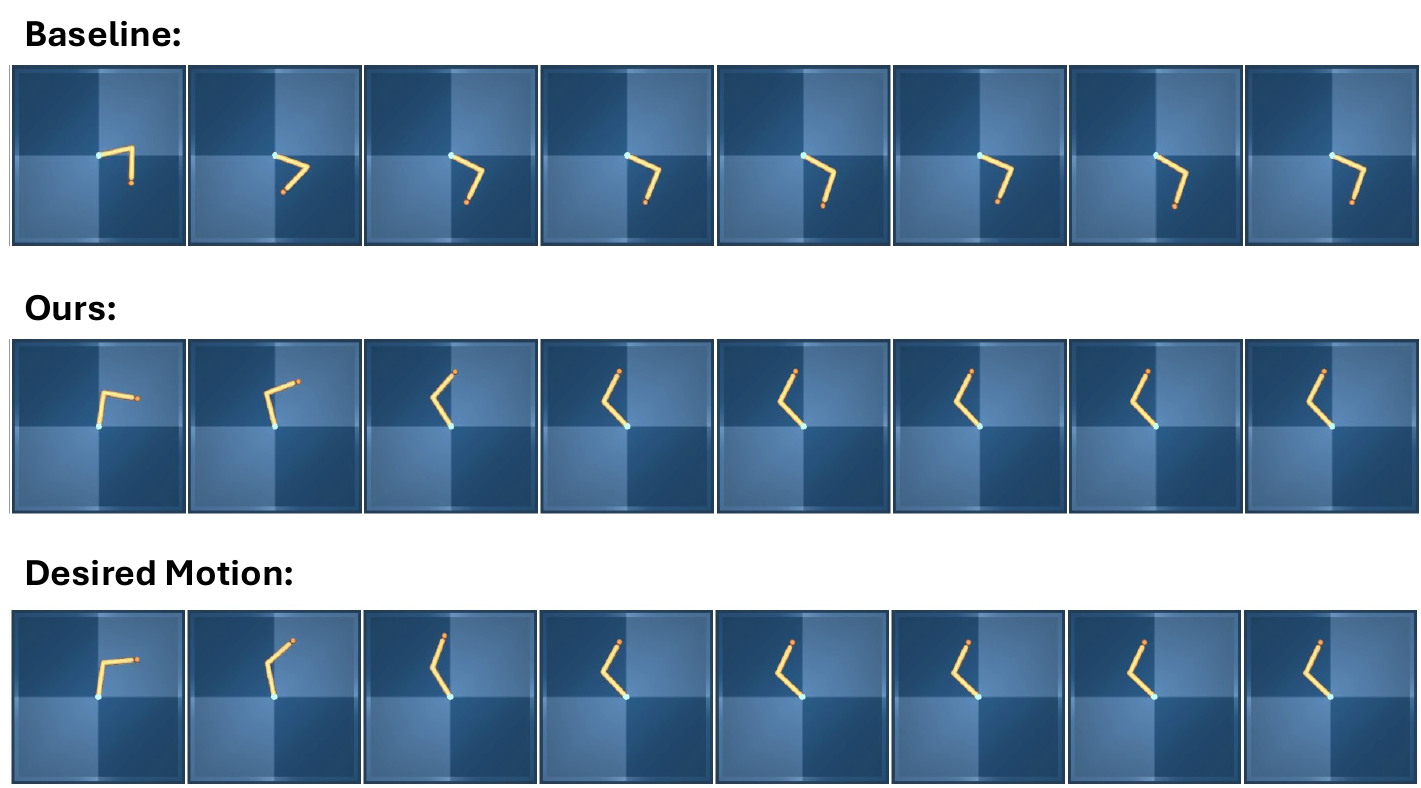}
\vspace{-3mm}
\caption{Visualized planning rollouts on the continuous control Reacher task.}
\label{fig:app_Reacher_visualization}
\vspace{2mm}
\includegraphics[width=0.92\textwidth]{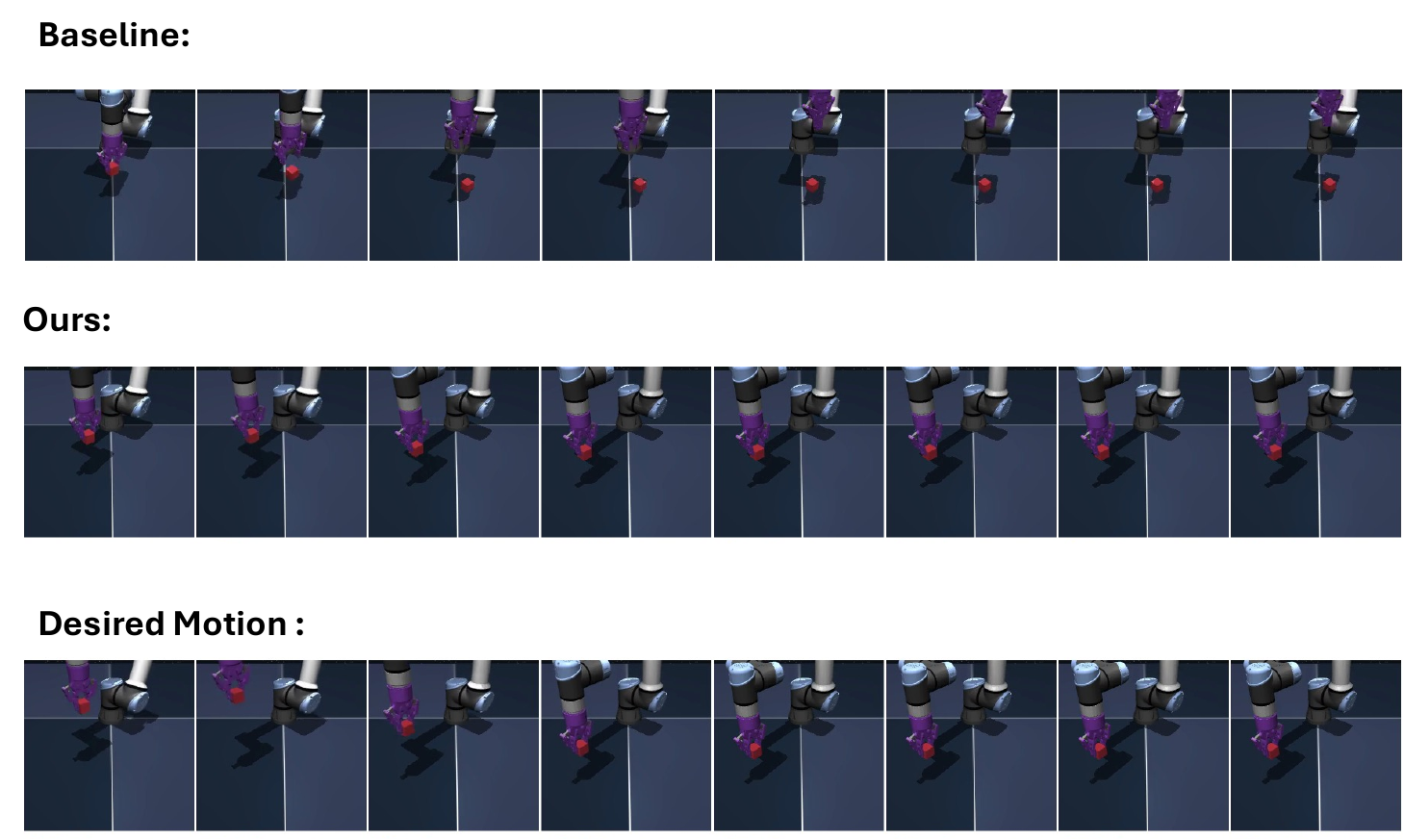}
\vspace{-3mm}
\caption{Multi-step predictor rollouts and manipulation trajectories on OGBench-Cube.}
\label{fig:app_arm_visualization}
\vspace{-2mm}
\end{figure*}

This section provides the per-configuration breakdowns for Section~\ref{sec:exp:planning}, qualitative rollout visualizations (\Cref{fig:app_PushT_visualization,fig:app_Tworoom_visualization,fig:app_Reacher_visualization,fig:app_arm_visualization}), per-backbone longer-horizon tables, and task-macro summaries.

\subsection{Temporal Straightening Breakdown by Configuration}
\Cref{tab:app-ts-dinov2_global,tab:app-ts-resnet_global,tab:app-ts-dinov2_spatial,tab:app-ts-resnet_spatial} reorganize the TS results by backbone and latent shape. U-Maze and Medium denote PointMaze--UMaze and PointMaze--Medium. Across all 32 comparisons, $\log\cosh$ improves over $u^2$, $\tfrac12 u^2$, $L_1$, and Smooth-$L_1$ (tying $\tfrac12 u^2$ once at $100.00\%$ on Wall under MPC with ResNet spatial latents).

\begin{table*}[!t]
\centering
\caption{TS planning success (\%) with DINOv2 global latents ($1\times384$). Mean $\pm$ std over three data-sampling seeds (50 episodes per seed). Best means are in \textbf{bold}.}
\vspace{-2mm}
\label{tab:app-ts-dinov2_global}
\setlength{\tabcolsep}{4.5pt}\renewcommand{\arraystretch}{1.06}
\resizebox{0.96\textwidth}{!}{%
\begin{tabular}{lcccccccccc}
\toprule
& \multicolumn{5}{c}{Open-loop} & \multicolumn{5}{c}{MPC} \\
\cmidrule(lr){2-6}\cmidrule(lr){7-11}
Task & $u^2$ & $\tfrac12u^2$ & $L_1$ ($|u|$) & Smooth-$L_1$ & Log-cosh & $u^2$ & $\tfrac12u^2$ & $L_1$ ($|u|$) & Smooth-$L_1$ & Log-cosh \\
\midrule
Wall & \appstat{36.67}{3.40} & \appstat{38.67}{2.49} & \appstat{40.67}{3.40} & \appstat{39.33}{2.49} & \appstat{\textbf{42.00}}{3.27} & \appstat{51.33}{4.11} & \appstat{54.00}{3.27} & \appstat{53.33}{4.99} & \appstat{55.33}{3.40} & \appstat{\textbf{56.67}}{4.11} \\
U-Maze & \appstat{30.67}{3.40} & \appstat{34.00}{3.27} & \appstat{36.00}{4.32} & \appstat{35.33}{2.49} & \appstat{\textbf{38.67}}{3.40} & \appstat{90.00}{3.27} & \appstat{92.67}{1.89} & \appstat{93.33}{2.49} & \appstat{94.67}{1.89} & \appstat{\textbf{96.00}}{0.00} \\
Medium & \appstat{16.67}{4.11} & \appstat{19.33}{4.11} & \appstat{18.67}{4.99} & \appstat{21.33}{4.11} & \appstat{\textbf{22.67}}{5.73} & \appstat{69.33}{3.40} & \appstat{73.33}{2.49} & \appstat{76.00}{3.27} & \appstat{74.67}{2.49} & \appstat{\textbf{78.00}}{2.83} \\
PushT & \appstat{2.67}{2.49} & \appstat{4.00}{1.63} & \appstat{4.67}{2.49} & \appstat{4.00}{1.63} & \appstat{\textbf{5.33}}{3.40} & \appstat{10.00}{1.63} & \appstat{12.00}{1.63} & \appstat{12.67}{2.49} & \appstat{13.33}{1.89} & \appstat{\textbf{14.67}}{0.94} \\
\bottomrule
\end{tabular}%
}
\vspace{3mm}
\caption{TS planning success (\%) with ResNet global latents ($1\times384$). Mean $\pm$ std over three data-sampling seeds (50 episodes per seed). Best means are in \textbf{bold}.}
\vspace{-2mm}
\label{tab:app-ts-resnet_global}
\resizebox{0.96\textwidth}{!}{%
\begin{tabular}{lcccccccccc}
\toprule
& \multicolumn{5}{c}{Open-loop} & \multicolumn{5}{c}{MPC} \\
\cmidrule(lr){2-6}\cmidrule(lr){7-11}
Task & $u^2$ & $\tfrac12u^2$ & $L_1$ ($|u|$) & Smooth-$L_1$ & Log-cosh & $u^2$ & $\tfrac12u^2$ & $L_1$ ($|u|$) & Smooth-$L_1$ & Log-cosh \\
\midrule
Wall & \appstat{78.67}{7.36} & \appstat{81.33}{5.73} & \appstat{82.67}{6.60} & \appstat{80.67}{5.73} & \appstat{\textbf{84.00}}{7.12} & \appstat{98.00}{1.63} & \appstat{99.33}{0.94} & \appstat{98.67}{1.89} & \appstat{99.33}{0.94} & \appstat{\textbf{100.00}}{0.00} \\
U-Maze & \appstat{42.00}{5.89} & \appstat{45.33}{5.73} & \appstat{46.67}{7.36} & \appstat{49.33}{5.73} & \appstat{\textbf{52.00}}{6.53} & \appstat{80.00}{3.27} & \appstat{83.33}{0.94} & \appstat{85.33}{2.49} & \appstat{84.00}{1.63} & \appstat{\textbf{86.67}}{0.94} \\
Medium & \appstat{44.67}{6.18} & \appstat{48.00}{6.53} & \appstat{51.33}{6.60} & \appstat{50.00}{5.89} & \appstat{\textbf{54.00}}{7.12} & \appstat{94.00}{1.63} & \appstat{96.00}{1.63} & \appstat{95.33}{2.49} & \appstat{96.67}{0.94} & \appstat{\textbf{98.00}}{0.00} \\
PushT & \appstat{14.67}{4.11} & \appstat{16.67}{4.11} & \appstat{16.00}{4.32} & \appstat{18.00}{3.27} & \appstat{\textbf{19.33}}{3.40} & \appstat{42.00}{6.53} & \appstat{45.33}{5.73} & \appstat{47.33}{5.73} & \appstat{46.00}{4.90} & \appstat{\textbf{48.67}}{4.99} \\
\bottomrule
\end{tabular}%
}
\vspace{-2mm}
\end{table*}

\begin{table*}[!t]
\centering
\caption{TS planning success (\%) with DINOv2 spatial latents ($14\times14\times8$). Mean $\pm$ std over three data-sampling seeds (50 episodes per seed). Best means are in \textbf{bold}.}
\vspace{-2mm}
\label{tab:app-ts-dinov2_spatial}
\setlength{\tabcolsep}{4.5pt}\renewcommand{\arraystretch}{1.06}
\resizebox{0.96\textwidth}{!}{%
\begin{tabular}{lcccccccccc}
\toprule
& \multicolumn{5}{c}{Open-loop} & \multicolumn{5}{c}{MPC} \\
\cmidrule(lr){2-6}\cmidrule(lr){7-11}
Task & $u^2$ & $\tfrac12u^2$ & $L_1$ ($|u|$) & Smooth-$L_1$ & Log-cosh & $u^2$ & $\tfrac12u^2$ & $L_1$ ($|u|$) & Smooth-$L_1$ & Log-cosh \\
\midrule
Wall & \appstat{86.00}{3.27} & \appstat{87.33}{2.49} & \appstat{86.67}{3.40} & \appstat{88.67}{2.49} & \appstat{\textbf{90.00}}{1.63} & \appstat{98.00}{1.63} & \appstat{99.33}{0.94} & \appstat{99.33}{0.94} & \appstat{98.67}{1.89} & \appstat{\textbf{100.00}}{0.00} \\
U-Maze & \appstat{82.00}{4.32} & \appstat{84.00}{1.63} & \appstat{89.33}{3.40} & \appstat{88.00}{2.83} & \appstat{\textbf{94.00}}{1.63} & \appstat{95.33}{2.49} & \appstat{98.00}{1.63} & \appstat{97.33}{2.49} & \appstat{98.67}{0.94} & \appstat{\textbf{100.00}}{0.00} \\
Medium & \appstat{72.00}{4.32} & \appstat{74.00}{4.32} & \appstat{76.67}{4.99} & \appstat{80.00}{3.27} & \appstat{\textbf{84.00}}{3.27} & \appstat{94.67}{2.49} & \appstat{96.67}{1.89} & \appstat{97.33}{1.89} & \appstat{96.00}{1.63} & \appstat{\textbf{98.67}}{0.94} \\
PushT & \appstat{78.00}{4.32} & \appstat{79.33}{3.40} & \appstat{80.67}{4.11} & \appstat{78.67}{3.40} & \appstat{\textbf{82.00}}{4.32} & \appstat{80.67}{2.49} & \appstat{82.67}{3.77} & \appstat{82.00}{4.32} & \appstat{84.00}{3.27} & \appstat{\textbf{85.33}}{4.99} \\
\bottomrule
\end{tabular}%
}
\vspace{3mm}
\caption{TS planning success (\%) with ResNet spatial latents ($14\times14\times8$). Mean $\pm$ std over three data-sampling seeds (50 episodes per seed). Best means are in \textbf{bold}.}
\vspace{-2mm}
\label{tab:app-ts-resnet_spatial}
\resizebox{0.96\textwidth}{!}{%
\begin{tabular}{lcccccccccc}
\toprule
& \multicolumn{5}{c}{Open-loop} & \multicolumn{5}{c}{MPC} \\
\cmidrule(lr){2-6}\cmidrule(lr){7-11}
Task & $u^2$ & $\tfrac12u^2$ & $L_1$ ($|u|$) & Smooth-$L_1$ & Log-cosh & $u^2$ & $\tfrac12u^2$ & $L_1$ ($|u|$) & Smooth-$L_1$ & Log-cosh \\
\midrule
Wall & \appstat{80.00}{3.27} & \appstat{82.00}{3.27} & \appstat{83.33}{3.40} & \appstat{81.33}{2.49} & \appstat{\textbf{84.67}}{2.49} & \appstat{98.67}{0.94} & \appstat{\textbf{100.00}}{0.00} & \appstat{98.67}{1.89} & \appstat{99.33}{0.94} & \appstat{\textbf{100.00}}{0.00} \\
U-Maze & \appstat{53.33}{6.60} & \appstat{57.33}{6.60} & \appstat{58.67}{7.36} & \appstat{61.33}{6.60} & \appstat{\textbf{64.67}}{8.38} & \appstat{94.67}{2.49} & \appstat{96.00}{1.63} & \appstat{97.33}{1.89} & \appstat{96.67}{2.49} & \appstat{\textbf{98.67}}{1.89} \\
Medium & \appstat{69.33}{4.11} & \appstat{73.33}{2.49} & \appstat{77.33}{3.40} & \appstat{75.33}{2.49} & \appstat{\textbf{80.67}}{0.94} & \appstat{96.00}{1.63} & \appstat{97.33}{0.94} & \appstat{96.67}{1.89} & \appstat{98.00}{1.63} & \appstat{\textbf{99.33}}{0.94} \\
PushT & \appstat{66.00}{3.27} & \appstat{68.00}{2.83} & \appstat{67.33}{3.40} & \appstat{69.33}{2.49} & \appstat{\textbf{70.67}}{0.94} & \appstat{84.00}{4.32} & \appstat{87.33}{3.77} & \appstat{89.33}{4.11} & \appstat{88.00}{3.27} & \appstat{\textbf{91.33}}{2.49} \\
\bottomrule
\end{tabular}%
}
\vspace{-2mm}
\end{table*}

\subsection{LeWorldModel and Longer-Horizon Breakdowns}
\Cref{tab:app-lewm} provides the task-by-mode breakdown for LeWM, and \Cref{tab:app-long-dino,tab:app-long-resnet} provide the per-backbone breakdowns for the longer-horizon evaluation in \Cref{tab:long_horizon}. Across all settings, $\log\cosh$ consistently outperforms $u^2$, $\tfrac12 u^2$, $L_1$, and Smooth-$L_1$.

\begin{table*}[!t]
\centering
\caption{LeWM planning success (\%). Mean $\pm$ std over three training seeds (50 episodes per seed). Best means are in \textbf{bold}.}
\vspace{-2mm}
\label{tab:app-lewm}
\setlength{\tabcolsep}{4.5pt}\renewcommand{\arraystretch}{1.06}
\resizebox{0.96\textwidth}{!}{%
\begin{tabular}{lcccccccccc}
\toprule
& \multicolumn{5}{c}{Open-loop} & \multicolumn{5}{c}{MPC} \\
\cmidrule(lr){2-6}\cmidrule(lr){7-11}
Task & $u^2$ & $\tfrac12u^2$ & $L_1$ ($|u|$) & Smooth-$L_1$ & Log-cosh & $u^2$ & $\tfrac12u^2$ & $L_1$ ($|u|$) & Smooth-$L_1$ & Log-cosh \\
\midrule
TwoRoom & \appstat{90.67}{2.49} & \appstat{91.33}{1.89} & \appstat{92.67}{2.49} & \appstat{92.00}{1.63} & \appstat{\textbf{94.67}}{1.89} & \appstat{96.67}{1.89} & \appstat{97.33}{1.89} & \appstat{96.67}{1.89} & \appstat{98.00}{1.63} & \appstat{\textbf{98.67}}{0.94} \\
PushT & \appstat{91.33}{2.49} & \appstat{90.67}{3.40} & \appstat{92.67}{2.49} & \appstat{92.00}{3.27} & \appstat{\textbf{93.33}}{2.49} & \appstat{96.67}{2.49} & \appstat{95.33}{2.49} & \appstat{96.00}{1.63} & \appstat{97.33}{1.89} & \appstat{\textbf{98.67}}{0.94} \\
Reacher & \appstat{80.67}{4.11} & \appstat{82.67}{3.40} & \appstat{84.00}{3.27} & \appstat{85.33}{2.49} & \appstat{\textbf{86.67}}{3.40} & \appstat{90.67}{3.40} & \appstat{92.67}{2.49} & \appstat{94.00}{1.63} & \appstat{93.33}{2.49} & \appstat{\textbf{95.33}}{1.89} \\
Cube & \appstat{70.67}{4.11} & \appstat{72.67}{3.77} & \appstat{72.00}{2.83} & \appstat{73.33}{2.49} & \appstat{\textbf{74.67}}{3.40} & \appstat{82.67}{4.11} & \appstat{84.67}{3.40} & \appstat{87.33}{3.40} & \appstat{86.00}{3.27} & \appstat{\textbf{88.67}}{2.49} \\
\bottomrule
\end{tabular}%
}
\vspace{3mm}
\caption{Longer-horizon success (\%) with DINOv2. Mean $\pm$ population std over three seeds (50 episodes per seed). Best means are in \textbf{bold}.}
\vspace{-2mm}
\label{tab:app-long-dino}
\resizebox{0.96\textwidth}{!}{%
\begin{tabular}{lcccccccccc}
\toprule
& \multicolumn{5}{c}{Open-loop} & \multicolumn{5}{c}{MPC} \\
\cmidrule(lr){2-6}\cmidrule(lr){7-11}
Task & $u^2$ & $\tfrac12u^2$ & $L_1$ ($|u|$) & Smooth-$L_1$ & Log-cosh & $u^2$ & $\tfrac12u^2$ & $L_1$ ($|u|$) & Smooth-$L_1$ & Log-cosh \\
\midrule
PushT & \appstat{10.67}{2.49} & \appstat{12.67}{3.40} & \appstat{13.33}{2.49} & \appstat{12.00}{3.27} & \appstat{\textbf{14.67}}{4.11} & \appstat{28.67}{6.60} & \appstat{30.67}{5.73} & \appstat{31.33}{5.73} & \appstat{32.67}{5.73} & \appstat{\textbf{33.33}}{5.73} \\
Medium & \appstat{64.67}{4.99} & \appstat{67.33}{5.73} & \appstat{69.33}{5.73} & \appstat{68.00}{4.90} & \appstat{\textbf{70.67}}{5.73} & \appstat{86.67}{4.11} & \appstat{89.33}{3.40} & \appstat{90.00}{3.27} & \appstat{91.33}{2.49} & \appstat{\textbf{92.67}}{3.40} \\
\bottomrule
\end{tabular}%
}
\vspace{3mm}
\caption{Longer-horizon success (\%) with ResNet. Mean $\pm$ population std over three seeds (50 episodes per seed). Best means are in \textbf{bold}.}
\vspace{-2mm}
\label{tab:app-long-resnet}
\resizebox{0.96\textwidth}{!}{%
\begin{tabular}{lcccccccccc}
\toprule
& \multicolumn{5}{c}{Open-loop} & \multicolumn{5}{c}{MPC} \\
\cmidrule(lr){2-6}\cmidrule(lr){7-11}
Task & $u^2$ & $\tfrac12u^2$ & $L_1$ ($|u|$) & Smooth-$L_1$ & Log-cosh & $u^2$ & $\tfrac12u^2$ & $L_1$ ($|u|$) & Smooth-$L_1$ & Log-cosh \\
\midrule
PushT & \appstat{8.67}{3.40} & \appstat{10.67}{4.11} & \appstat{10.00}{3.27} & \appstat{11.33}{4.11} & \appstat{\textbf{12.67}}{5.73} & \appstat{30.67}{7.72} & \appstat{33.33}{6.80} & \appstat{35.33}{5.73} & \appstat{34.00}{6.53} & \appstat{\textbf{36.67}}{6.60} \\
Medium & \appstat{62.67}{6.60} & \appstat{65.33}{5.25} & \appstat{66.00}{4.90} & \appstat{67.33}{5.73} & \appstat{\textbf{68.67}}{6.80} & \appstat{87.33}{3.77} & \appstat{90.67}{4.11} & \appstat{92.00}{3.27} & \appstat{91.33}{3.40} & \appstat{\textbf{93.33}}{2.49} \\
\bottomrule
\end{tabular}%
}
\vspace{-2mm}
\end{table*}

\subsection{Task-Macro Summaries and Regularizer Ablation Protocols}
\label{app:ablations}
\Cref{tab:app-macro} summarizes the four standard-horizon tasks within each configuration. Notably, the DINOv2 spatial configuration achieves an average open-loop gain of $8.00$ percentage points over $u^2$ and $6.33$ percentage points over $\tfrac12 u^2$.

For the representation-regularizer ablations reported in Section~\ref{sec:exp:planning}, the SIGReg sweep in \Cref{tab:app-sigreg-sweep} varies $k \in \{0, 0.03, 0.09, 0.27, 0.50, 0.81\}$ on TwoRoom with $\log\cosh$ fixed, evaluating each trained model over 50 CEM planning episodes. In the TS curvature ablation, models are trained with $\log\cosh$ either with the native temporal-straightening weight enabled or with $\lambda = 0$, yielding open-loop gains of $+56$ percentage points on Wall and $+78$ percentage points on U-Maze when regularization is active.

\begin{table}[!t]
\centering
\caption{Equal-task macro success rates (\%) and $\log\cosh$ gains (percentage points, pp). Each row averages the four task means within a configuration; OL denotes open-loop execution.}
\vspace{-2mm}
\label{tab:app-macro}
\setlength{\tabcolsep}{4pt}\renewcommand{\arraystretch}{1.06}
\resizebox{\columnwidth}{!}{%
\begin{tabular}{llrrrrr}
\toprule
& & \multicolumn{3}{c}{Macro success (\%)} & \multicolumn{2}{c}{Log-cosh gain (pp)} \\
\cmidrule(lr){3-5}\cmidrule(lr){6-7}
Configuration & Mode & $u^2$ & $\tfrac12u^2$ & Log-cosh & vs.\ $u^2$ & vs.\ $\tfrac12u^2$ \\
\midrule
DINOv2, $1\times384$ & OL & 21.67 & 24.00 & 27.17 & 5.50 & 3.17 \\
DINOv2, $1\times384$ & MPC & 55.17 & 58.00 & 61.34 & 6.17 & 3.34 \\
ResNet, $1\times384$ & OL & 45.00 & 47.83 & 52.33 & 7.33 & 4.50 \\
ResNet, $1\times384$ & MPC & 78.50 & 81.00 & 83.34 & 4.84 & 2.34 \\
DINOv2, $14\times14\times8$ & OL & 79.50 & 81.17 & 87.50 & 8.00 & 6.33 \\
DINOv2, $14\times14\times8$ & MPC & 92.17 & 94.17 & 96.00 & 3.83 & 1.83 \\
ResNet, $14\times14\times8$ & OL & 67.17 & 70.17 & 75.17 & 8.00 & 5.00 \\
ResNet, $14\times14\times8$ & MPC & 93.34 & 95.17 & 97.33 & 3.99 & 2.16 \\
LeWM & OL & 83.34 & 84.34 & 87.34 & 4.00 & 3.00 \\
LeWM & MPC & 91.67 & 92.50 & 95.34 & 3.67 & 2.84 \\
\bottomrule
\end{tabular}%
}
\vspace{-2mm}
\end{table}

\section{Autonomous Driving Experimental Details}
\label{app:driving-details}

We evaluate bounded predictive regulation on NAVSIM~\citep{dauner2024navsim} across two model scales and protocols: (1) a convolutional JEPA model (HENet-TinyM) evaluated on perception MAE and open-loop trajectory errors (ADE/FDE), and (2) a V-JEPA~2.1 ViT-L/16 backbone paired with a Drive-JEPA perception-based planner evaluated on closed-loop PDMS.

\subsection{Convolutional JEPA (HENet-TinyM) Protocol}

\paragraph{Dataset and clip construction.}
Self-supervised training uses multi-camera logs from NAVSIM \texttt{trainval} (OpenScene/nuPlan) without human annotations or \texttt{navtest} scenes. The split contains 1,310 logs, 19,376 scenes, and 723,019 keyframes sampled at $2\,\mathrm{Hz}$ across eight camera views. Within each scene, we build non-overlapping single-camera clips of $T=8$ frames separated by $0.5\,\mathrm{s}$ ($\pm 0.15\,\mathrm{s}$ tolerance, $3.5\,\mathrm{s}$ span), yielding 708,416 clips across 18,920 scenes (5,667,328 images). All cameras share one encoder. Frames are resized bilinearly to $1024\times512$ with ImageNet normalization.

\paragraph{Architecture and tube masking.}
The model uses a HENet-TinyM~\citep{xia2024henet} context encoder (7.6M parameters, four stages with channels $64/128/192/384$ and block counts $4/3/8/6$), an EMA target encoder, and a 3D convolutional predictor (0.77M parameters, three blocks of $3\times3\times3$ depthwise and $1\times1\times1$ pointwise convolutions with channel-wise LayerNorm and GELU). An 8-frame clip at stride 32 produces an $8\times16\times32$ latent grid. For each clip, four spatial rectangles (each covering $15\%$--$20\%$ of the grid, aspect ratio in $[0.75, 1.5]$) are repeated across all eight frames (masking $43\%$--$49\%$ of tokens). Masked input pixels are zeroed, intermediate features are re-masked, and Sparse BatchNorm computes statistics only over visible tokens. Target parameters update via $\theta_{\mathrm{tgt}} \leftarrow m\theta_{\mathrm{tgt}} + (1-m)\theta_{\mathrm{ctx}}$, with $m$ annealed linearly from $0.996$ to $1.0$.

\begin{figure}[!t]
\centering
\includegraphics[width=\columnwidth]{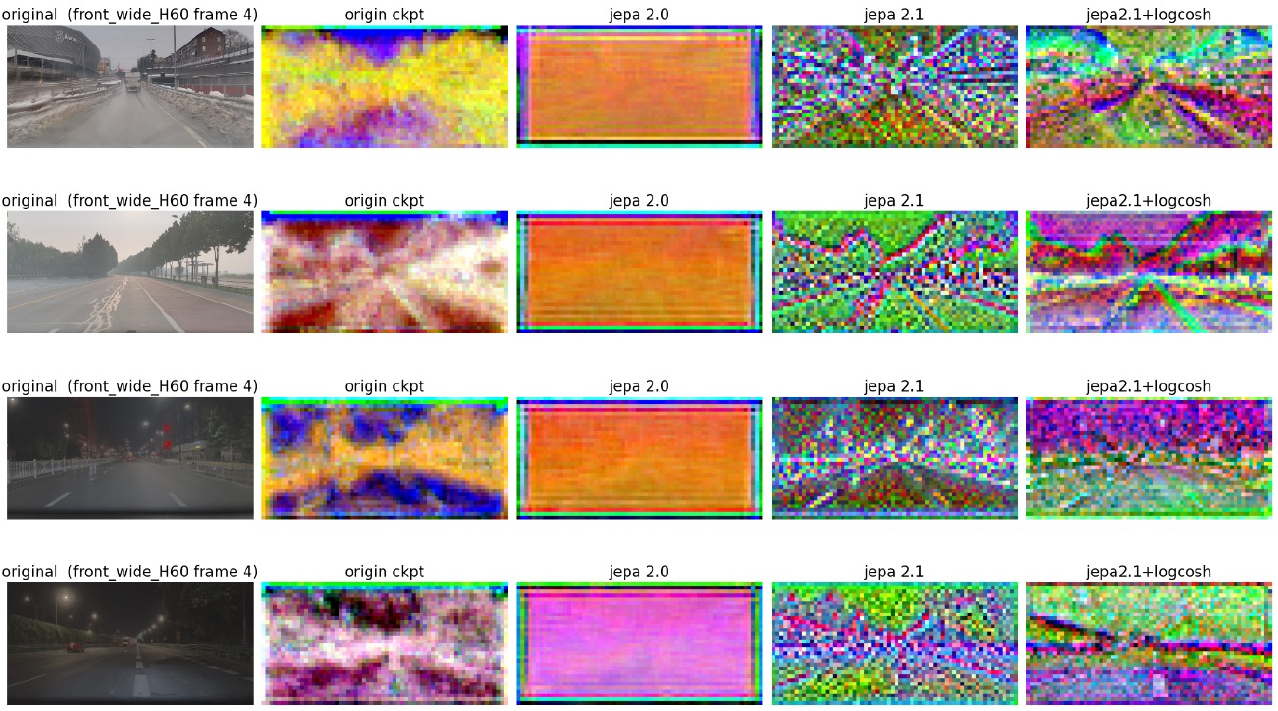}
\vspace{-4mm}
\caption{Additional latent feature visualizations on NAVSIM driving scenes comparing $u^2$ (Baseline) and $\log\cosh$ (Ours).}
\label{fig:app_feature_visualization}
\vspace{-2mm}
\end{figure}

\paragraph{Two-branch objective and optimization.}
Let $h$ denote target features and $z_b$ denote the output of predictor head $b \in \{\mathrm{pred}, \mathrm{ctx}\}$ for masked prediction and visible context reconstruction. At cell $i$, vectors are normalized across $C=384$ channels via affine-free LayerNorm ($\operatorname{LN}$):
\begin{equation}
d_b(i)=\frac{1}{C}\sum_{c=1}^{C}
\rho\!\left(\operatorname{LN}(z_b)_{c,i}-\operatorname{LN}(h)_{c,i}\right).
\label{eq:driving-cell-distance}
\end{equation}
Over masked cells $\mathcal M$ and visible cells $\mathcal V$, the branch losses are
\begin{equation}
\begin{aligned}
\mathcal L_{\mathrm{pred}}&=\frac{1}{|\mathcal M|}\sum_{i\in\mathcal M}d_{\mathrm{pred}}(i), \\
\mathcal L_{\mathrm{ctx}}&=\frac{1}{|\mathcal V|}\sum_{i\in\mathcal V}w_i d_{\mathrm{ctx}}(i),
\end{aligned}
\label{eq:driving-branch-losses}
\end{equation}
with boundary distance weights
\begin{equation}
w_i=\left[\max\!\left(1,\min_{j\in\mathcal M}\|p_i-p_j\|_2\right)\right]^{-1/2},
\quad p_i=(t_i,y_i,x_i).
\label{eq:driving-context-weights}
\end{equation}
The total loss is $\mathcal L_{\mathrm{pred}} + \lambda\mathcal L_{\mathrm{ctx}}$, where $\lambda$ ramps linearly from $0$ to $0.05$ over 20,000 updates. We train for 10 epochs (110,690 steps) using AdamW ($\beta=(0.9, 0.95)$, peak lr $6\times 10^{-4}$ with 1-epoch warmup and cosine decay, cosine weight decay $0.04 \to 0.4$, grad clip $1.0$) on 16 A100 GPUs (global batch 64 clips). \Cref{tab:app-driving-full,tab:app-driving-gains} report the complete metrics and relative reductions; additional feature maps are shown in \Cref{fig:app_feature_visualization}.

\begin{table}[!t]
\centering
\caption{Complete HENet-TinyM driving results on NAVSIM. Lower is better for all metrics.}
\vspace{-2mm}
\label{tab:app-driving-full}
\setlength{\tabcolsep}{3.2pt}\renewcommand{\arraystretch}{1.06}
\resizebox{\columnwidth}{!}{%
\begin{tabular}{llrrrrrrr}
\toprule
& & \multicolumn{5}{c}{Perception: MAE $\downarrow$} & \multicolumn{2}{c}{Planning $\downarrow$} \\
\cmidrule(lr){3-7}\cmidrule(lr){8-9}
Init. & Loss & Veh. & Ped. & Bic. & Cone & Barrier & ADE & FDE \\
\midrule
Pretrained & $u^2$ & 2.93 & 3.79 & 0.053 & 1.02 & 0.158 & 7.91 & 14.21 \\
Pretrained & $\tfrac12u^2$ & 2.94 & 3.81 & 0.053 & 1.02 & 0.158 & 7.95 & 14.28 \\
Pretrained & Log-cosh & \textbf{2.65} & \textbf{3.43} & \textbf{0.048} & \textbf{0.91} & \textbf{0.133} & \textbf{7.15} & \textbf{12.85} \\
\midrule
Scratch & $u^2$ & 3.08 & 3.98 & 0.055 & 1.07 & 0.165 & 8.36 & 15.16 \\
Scratch & $\tfrac12u^2$ & 3.09 & 4.00 & 0.055 & 1.07 & 0.166 & 8.40 & 15.24 \\
Scratch & Log-cosh & \textbf{2.71} & \textbf{3.50} & \textbf{0.049} & \textbf{0.93} & \textbf{0.136} & \textbf{7.30} & \textbf{13.12} \\
\bottomrule
\end{tabular}%
}
\vspace{-2mm}
\end{table}

\begin{table}[!t]
\centering
\small
\caption{Relative error reductions (\%) of $\log\cosh$ on the HENet-TinyM driving benchmark.}
\vspace{-2mm}
\label{tab:app-driving-gains}
\setlength{\tabcolsep}{5pt}\renewcommand{\arraystretch}{1.06}
\resizebox{\columnwidth}{!}{%
\begin{tabular}{lrrrr}
\toprule
& \multicolumn{2}{c}{Pretrained} & \multicolumn{2}{c}{From scratch} \\
\cmidrule(lr){2-3}\cmidrule(lr){4-5}
Downstream Metric & vs.\ $u^2$ & vs.\ $\tfrac12u^2$ & vs.\ $u^2$ & vs.\ $\tfrac12u^2$ \\
\midrule
Vehicle MAE & 9.6 & 9.9 & 12.0 & 12.3 \\
Pedestrian MAE & 9.5 & 10.0 & 12.1 & 12.5 \\
Bicycle MAE & 9.4 & 9.4 & 10.9 & 10.9 \\
Traffic-cone MAE & 10.8 & 10.8 & 13.1 & 13.1 \\
Barrier MAE & 15.8 & 15.8 & 17.6 & 18.1 \\
Planning ADE (m) & 9.6 & 10.1 & 12.7 & 13.1 \\
Planning FDE (m) & 9.6 & 10.0 & 13.5 & 13.9 \\
\bottomrule
\end{tabular}%
}
\vspace{-2mm}
\end{table}

\subsection{V-JEPA~2.1 ViT-L and Closed-Loop PDMS Protocol}
\label{app:vjepa21-pb-details}

\paragraph{V-JEPA~2.1 ViT-L/16 pretraining.}
We pretrain a V-JEPA~2.1~\citep{mur2026v} ViT-L/16 encoder (with RoPE) and a 24-layer predictor (width 384) using 4-layer hierarchical target distillation (\texttt{predict\_all} with distance-weighted context loss). The baseline MSE objective is natively $\tfrac12\|z - h\|_2^2$ (\texttt{loss\_exp=2}), which we compare against element-wise $\log\cosh(z - h)$. Pretraining uses all 8 cameras from NAVSIM \texttt{trainval} as independent 8-frame clips at $2\,\mathrm{Hz}$ ($3.5\,\mathrm{s}$ span, tubelet size 2, $256\times448$ resolution with random resized crop and horizontal flip) under 3D multi-block masking (8 blocks at scale $0.15$ and 2 blocks at scale $0.70$). Models train in FP32 with AdamW (lr $2\times 10^{-5} \to 1\times 10^{-4}$ over a 10-epoch warmup, weight decay $0.04$, EMA $0.99925$) across 40 GPUs (batch size 16/GPU, global batch 640, 1,000 steps/epoch). Both encoders are evaluated using epoch-9 pretraining weights.

\paragraph{Perception-based planner architecture.}
We attach the Drive-JEPA perception-based planning head to the pretrained ViT-L EMA target encoder. Given two consecutive front-camera (\texttt{CAM\_F0}) frames $(t-1, t)$ (one tubelet, cropped by 28 rows top/bottom and resized to $512\times256$, with training GridMask), the encoder outputs $16\times32 = 512$ tokens of width 1024, projected linearly to 256D. Ego status (4D command, 2D velocity, 2D acceleration) is added to $32 \times 8 = 256$ learnable queries representing 32 proposals over 8 future poses ($4.0\,\mathrm{s}$ horizon at $0.5\,\mathrm{s}$ steps). A shared Refiner updates queries over 4 iterations: an MLP decodes poses $(x, y, \text{heading})$, followed by BEVFormer-style deformable self-attention across the $32\times 8$ grid, spatial cross-attention into visual tokens (projecting the 4 ego corners across 4 heights via \texttt{lidar2img}), and an FFN. A Scorer MLP max-pools across the 8 poses per proposal to predict 6 PDM sub-score logits, selecting the top-scoring proposal at inference.

\paragraph{Fine-tuning and evaluation.}
On \texttt{navtrain} (84,792 train / 18,136 val samples), the full model is fine-tuned for 20 epochs (6,620 steps, 331 steps/epoch) across 64 GPUs (batch size 4/GPU, global batch 256) in FP32 using AdamW (head lr $1\times 10^{-4}$, backbone lr $1\times 10^{-5}$, 1-epoch warmup then cosine decay to 0, grad clip $1.0$) with the loss weights in \Cref{tab:app-pb-losses}. Checkpoints are selected by validation PDMS on \texttt{navval} without accessing \texttt{navtest}. Closed-loop evaluation covers all 12,146 scenes of NAVSIM v1 \texttt{navtest} (0 failed scenes), as detailed in \Cref{tab:app-driving-pdms-full}.

\begin{table*}[ht]
\centering
\small
\caption{Supervision terms and loss weights for end-to-end fine-tuning of the perception-based planner on \texttt{navtrain}.}
\vspace{-2mm}
\label{tab:app-pb-losses}
\setlength{\tabcolsep}{6pt}\renewcommand{\arraystretch}{1.06}
\begin{tabular}{lp{0.68\textwidth}r}
\toprule
Loss Term & Description & Weight \\
\midrule
Trajectory min-$L_1$ & Minimum $L_1$ distance between the 32 refined proposals and the ground-truth trajectory across the 4 refinement stages (earlier stages decayed by $0.1$). & $1.0$ \\
Pseudo-target min-$L_1$ & Minimum $L_1$ supervision against 4 high-PDMS anchor trajectories sampled per scene from an 8,192-trajectory anchor vocabulary. & $0.5$ \\
Online PDMS Scorer & Binary cross-entropy (BCE) supervising the Scorer MLP against true PDM sub-scores evaluated online via the \texttt{navtrain} metric cache. & $1.0$ \\
Key Agent Prediction & Auxiliary validity BCE and state $L_1$ regression for surrounding traffic agents. & $1.0 / 0.1$ \\
Ego-Area Prediction & Auxiliary BCE loss for ego-vehicle drivable area occupancy. & $2.0$ \\
\bottomrule
\end{tabular}
\vspace{3mm}
\caption{Complete epoch-by-epoch closed-loop evaluation on NAVSIM \texttt{navtest} (12,146 scenes) comparing V-JEPA~2.1 ViT-L encoders pretrained with matched MSE ($\tfrac12 u^2$) vs.\ $\log\cosh$. Best values at each stage are in \textbf{bold}.}
\vspace{-2mm}
\label{tab:app-driving-pdms-full}
\resizebox{0.92\textwidth}{!}{%
\begin{tabular}{llrrrrrrr}
\toprule
Fine-tuning Stage & Pretraining Loss & NC $\uparrow$ & DAC $\uparrow$ & DDC $\uparrow$ & TTC $\uparrow$ & Comf. $\uparrow$ & EP $\uparrow$ & \textbf{PDMS} $\uparrow$ \\
\midrule
Epoch 0 (Step 331) & Matched MSE ($\tfrac12 u^2$) & 96.5 & 89.5 & \textbf{97.8} & 90.4 & 76.9 & 58.2 & 69.03 \\
Epoch 0 (Step 331) & Log-cosh & \textbf{97.0} & \textbf{95.2} & 97.7 & \textbf{91.6} & \textbf{77.2} & \textbf{65.2} & \textbf{75.35} \\
\midrule
Epoch 3 (Step 1,324) & Matched MSE ($\tfrac12 u^2$) & \textbf{98.6} & 93.6 & \textbf{96.9} & 93.1 & \textbf{98.6} & 77.9 & 83.87 \\
Epoch 3 (Step 1,324) & Log-cosh & 98.1 & \textbf{94.5} & 96.5 & \textbf{93.7} & 97.5 & \textbf{80.7} & \textbf{85.58} \\
\midrule
Epoch 5 (Step 1,986) & Matched MSE ($\tfrac12 u^2$) & \textbf{98.7} & \textbf{96.7} & \textbf{96.8} & \textbf{95.6} & \textbf{99.8} & 82.2 & 88.66 \\
Epoch 5 (Step 1,986) & Log-cosh & 98.6 & 96.6 & 96.5 & 95.3 & 98.2 & \textbf{83.5} & \textbf{88.74} \\
\midrule
Epoch 7 (Step 2,648) & Matched MSE ($\tfrac12 u^2$) & 98.6 & 94.8 & \textbf{96.5} & 94.9 & \textbf{99.9} & 82.0 & 87.21 \\
Epoch 7 (Step 2,648) & Log-cosh & \textbf{98.7} & \textbf{97.2} & 96.1 & \textbf{95.9} & 99.8 & \textbf{83.9} & \textbf{89.77} \\
\midrule
Best Val (Step 4,965, Ep.\ 14) & Matched MSE ($\tfrac12 u^2$) & 98.43 & 96.95 & \textbf{95.54} & 94.90 & 99.78 & 86.96 & 90.48 \\
Best Val (Step 4,634, Ep.\ 13) & Log-cosh & \textbf{98.53} & \textbf{97.46} & 95.14 & \textbf{95.06} & \textbf{99.86} & \textbf{87.19} & \textbf{90.91} \\
\bottomrule
\end{tabular}%
}
\vspace{-2mm}
\end{table*}

\section{Physical-State Probing and Visualization Details}
\label{app:visualizations}

\subsection{Physical-State Probe Protocol}
\label{app:probe-details}
The PushT probing evaluation trains linear and MLP readouts on frozen LeWM encoders to predict agent position, block position, and block orientation. Using a fixed train/val/test split, spatial coordinates are scaled to $[-1, 1]$ and orientation $\theta$ is encoded as $(\cos\theta, \sin\theta)$. Probes are trained with MSE using Adam and validation early stopping across three probe initialization seeds. \Cref{tab:app-probe-linear,tab:app-probe-mlp,tab:app-probe-gains} report the complete results.

\begin{table}
\centering
\caption{Linear probe results on PushT with frozen LeWM encoders (mean $\pm$ std across three probe seeds).}
\vspace{-2mm}
\label{tab:app-probe-linear}
\setlength{\tabcolsep}{4pt}\renewcommand{\arraystretch}{1.06}
\resizebox{\columnwidth}{!}{%
\begin{tabular}{lrrrrrr}
\toprule
& \multicolumn{2}{c}{$u^2$} & \multicolumn{2}{c}{$\tfrac12u^2$} & \multicolumn{2}{c}{Log-cosh} \\
\cmidrule(lr){2-3}\cmidrule(lr){4-5}\cmidrule(lr){6-7}
Target Quantity & MSE $\downarrow$ & $r\uparrow$ & MSE $\downarrow$ & $r\uparrow$ & MSE $\downarrow$ & $r\uparrow$ \\
\midrule
Agent location & \appstat{0.0513}{0.0051} & 0.9743 & \appstat{0.0497}{0.0047} & 0.9751 & \appstat{\textbf{0.0383}}{0.0039} & \textbf{0.9817} \\
Block location & \appstat{0.0287}{0.0035} & 0.9857 & \appstat{0.0263}{0.0033} & 0.9871 & \appstat{\textbf{0.0203}}{0.0027} & \textbf{0.9903} \\
Block orientation & \appstat{0.1823}{0.0173} & 0.9053 & \appstat{0.1733}{0.0167} & 0.9097 & \appstat{\textbf{0.1373}}{0.0137} & \textbf{0.9277} \\
\bottomrule
\end{tabular}%
}
\vspace{2mm}
\caption{MLP probe results on PushT with frozen LeWM encoders (mean $\pm$ std across three probe seeds).}
\vspace{-2mm}
\label{tab:app-probe-mlp}
\resizebox{\columnwidth}{!}{%
\begin{tabular}{lrrrrrr}
\toprule
& \multicolumn{2}{c}{$u^2$} & \multicolumn{2}{c}{$\tfrac12u^2$} & \multicolumn{2}{c}{Log-cosh} \\
\cmidrule(lr){2-3}\cmidrule(lr){4-5}\cmidrule(lr){6-7}
Target Quantity & MSE $\downarrow$ & $r\uparrow$ & MSE $\downarrow$ & $r\uparrow$ & MSE $\downarrow$ & $r\uparrow$ \\
\midrule
Agent location & \appstat{0.0041}{0.0005} & 0.9979 & \appstat{0.0038}{0.0004} & 0.9981 & \appstat{\textbf{0.0024}}{0.0003} & \textbf{0.9993} \\
Block location & \appstat{0.0011}{0.0002} & 0.9991 & \appstat{0.0010}{0.0002} & 0.9992 & \appstat{\textbf{0.0007}}{0.0001} & \textbf{0.9995} \\
Block orientation & \appstat{0.0207}{0.0021} & 0.9897 & \appstat{0.0193}{0.0023} & 0.9907 & \appstat{\textbf{0.0143}}{0.0017} & \textbf{0.9933} \\
\bottomrule
\end{tabular}%
}
\vspace{2mm}
\caption{Relative probe MSE reductions (\%) achieved by $\log\cosh$ over $u^2$ and $\tfrac12 u^2$ on PushT.}
\vspace{-2mm}
\label{tab:app-probe-gains}
\setlength{\tabcolsep}{6pt}
\begin{tabular}{llrr}
\toprule
Target Quantity & Readout Type & vs.\ $u^2$ & vs.\ $\tfrac12u^2$ \\
\midrule
Agent location & Linear & 25.3 & 22.9 \\
Agent location & MLP & 41.5 & 36.8 \\
Block location & Linear & 29.3 & 22.8 \\
Block location & MLP & 36.4 & 30.0 \\
Block orientation & Linear & 24.7 & 20.8 \\
Block orientation & MLP & 30.9 & 25.9 \\
\bottomrule
\end{tabular}
\vspace{-2mm}
\end{table}

\subsection{Scope of Geometric Visualizations}
\label{app:feature-visualization}
In Section~\ref{sec:exp:visualization}, the PointMaze goal-distance heatmaps (\Cref{fig:latent_distance}) evaluate static encoder representations by plotting squared Euclidean distances $\|f_\theta(o) - f_\theta(o_{\mathrm{goal}})\|_2^2$ across grid states without predictor rollouts. The action-space plots (\Cref{fig:action_optimization,fig:action_cost_contours}) evaluate the multi-step planning objective $\mathcal J(a) = \|\hat Z(a) - Z_{\mathrm{goal}}\|_2^2$ over a 2D action subspace $(a_1, a_2)$, comparing learned cost landscapes against ground-truth simulator dynamics.

\section{Compute and Implementation Cost}
\label{app:reproducibility}
Evaluating element-wise $\log\cosh$ via Equation~\ref{eq:app-analytic-stable-logcosh} introduces zero additional parameters and no extra forward or backward passes. \Cref{tab:app-compute} summarizes the compute resources for the NAVSIM self-supervised pretraining and downstream planner fine-tuning stages.

\begin{table*}
\centering
\small
\setlength{\tabcolsep}{8pt}\renewcommand{\arraystretch}{1.08}
\caption{Compute specifications for the NAVSIM self-supervised pretraining (SSL) and downstream planner fine-tuning (FT) stages.}
\vspace{-2mm}
\label{tab:app-compute}
\begin{tabular}{lll}
\toprule
Quantity & HENet-TinyM (SSL) & V-JEPA~2.1 ViT-L (SSL / Planner FT) \\
\midrule
Accelerators & 16 $\times$ A100 80\,GB & 40 GPUs (SSL) / 64 GPUs (FT) \\
Numerical precision & FP32 & FP32 \\
Global batch size & 64 clips (512 frames) & 640 clips (SSL) / 256 samples (FT) \\
Training duration & 10 epochs (110,690 steps) & 9 epochs (9,000 steps SSL) / 20 epochs (6,620 steps FT) \\
Wall-clock time (SSL) & $\approx 25\,\mathrm{h}$ ($\approx 400$ GPU-hours) & Identical across $\tfrac12 u^2$ and $\log\cosh$ \\
\bottomrule
\end{tabular}
\vspace{-2mm}
\end{table*}

\end{document}